%% file: main.tex
\documentclass[final,3p,times]{elsarticle} 

\journal{arXiv preprint}

\usepackage[dvipsnames,table]{xcolor}
\usepackage{amssymb,amsmath,amsthm}
\usepackage{lipsum,hyperref,makecell,tabularx,multirow,booktabs,soul,bbm,lipsum,cleveref, natbib,ulem,fontawesome5,xurl,microtype}
\usepackage{lineno} % single column
\usepackage[figuresleft]{rotating}

\usepackage{geometry}
\begin{document}

\input{src/functions.tex}
\input{src/frontmatter_els.tex}

%% main text
% \linenumbers
\input{src/body_main.tex}

\section*{CRediT authorship contribution statement}
\textbf{Hyeonbeen Lee:} Conceptualization, Data curation, Formal analysis, Investigation, Methodology, Software, Validation, Visualization, Writing -- original draft. \textbf{Min-Jae Jung:} Conceptualization, Formal analysis, Investigation, Resources, Writing -- review \& editing.
\textbf{Tae-Kyeong Yeu:} Data curation, Project administration, Resources, Writing -- review \& editing.
\textbf{Jong-Boo Han:} Data curation, Project administration, Resources, Writing -- review \& editing.
\textbf{Daegil Park:} Data curation, Project administration, Resources, Validation, Writing -- review \& editing.
\textbf{Simon Stepputtis:} Resources, Validation, Supervision, Writing -- review \& editing.
\textbf{Jin-Gyun Kim:} Funding acquisition, Project administration, Resources, Validation, Supervision, Writing -- review \& editing.

\section*{Declaration of competing interest}
The authors declare that they have no known competing financial interests or personal relationships that could have appeared to influence the work reported in this paper.

\section*{Declaration of generative AI and AI-assisted technologies in the manuscript preparation process}
During the preparation of this work the author(s) used Anthropic Claude in order to improve the clarity and readability of the text. After using this tool/service, the author(s) reviewed and edited the content as needed and take(s) full responsibility for the content of the published article.

\section*{Acknowledgements}
This research was supported by a grant from Endowment Project of ``Development of Core Technologies for Operation of Marine Robots based on Cyber-Physical System'' funded by Korea Research Institute of Ships and Ocean Engineering (PES5200).

\section*{Data availability}
Code is publicly available at \github. Data and model checkpoints are archived on Zenodo~\citep{lee_2026_23026021}.

\bibliographystyle{elsarticle-num-names}
\bibliography{src/references}

\end{document}

%% file: src/functions.tex
\newcommand{\mycellleft}[1]{\multicolumn{2}{c|}{\makecell[c]{#1}}}
\newcommand{\mycellleftdoublerow}[1]{\multicolumn{2}{c}{\multirow{2}{*}{\makecell[c]{#1}}}}
\newcommand{\mycell}[1]{\multicolumn{2}{c|}{\makecell[c]{#1}}}
\newcommand{\mycellplain}[1]{\multicolumn{2}{c}{\makecell[c]{#1}}}

\newcommand{\red}[1]{{\color{red}{#1}}}
\newcommand{\blue}[1]{{\color{blue}{#1}}}
\newcommand{\green}[1]{{\color{LimeGreen}{#1}}}

\newcommand{\github}{\url{https://github.com/leehyeonbeen/FDN}}

%% file: src/frontmatter_els.tex
\begin{frontmatter}

    \title{Frequency-aware~decomposition~learning for~sensorless wrench~estimation in~vibration-rich~robotic~contact} %% Article title

    \author[vt,khu]{Hyeonbeen~Lee} %% Author name
    \ead{leehyeonbeen@vt.edu}

    \author[khu]{Min-Jae~Jung}
    \ead{jmhahh@khu.ac.kr}

    \author[kriso]{Tae-Kyeong~Yeu}
    \ead{yeutk@kriso.re.kr}

    \author[kriso]{Jong-Boo~Han}
    \ead{jbhan@kriso.re.kr}

    \author[kriso]{Daegil~Park}
    \ead{daegilpark@kriso.re.kr}

    \author[vt]{Simon~Stepputtis}
    \ead{stepputtis@vt.edu}

    \author[khu]{Jin-Gyun~Kim\corref{cor1}}
    \ead{jingyun.kim@khu.ac.kr}
    \cortext[cor1]{Corresponding author}

    %% Author affiliation
    \affiliation[vt]{organization={Department of Mechanical Engineering, Virginia Tech},% Department and Organization
        % city={Blacksburg},
        postcode={24061},
        state={VA},
        country={United States}}

    %% Author affiliation
    \affiliation[khu]{organization={School of Mechanical Engineering, Kyung Hee University},% Department and Organization
        postcode={17104},
        state={Gyeonggi-do},
        country={South Korea}}

    %% Author affiliation
    \affiliation[kriso]{organization={Korea Research Institute of Ships and Ocean Engineering},% Department and Organization
        postcode={34103},
        state={Daejeon},
        country={South Korea}}
    %%%%%%%%%%%%%%%%%%%%%%%%%%%%%%%%%%%%%%%%%%%%%%%%%%%%%%%%%%%%%%%
    %%%%%%%%%%%%%%%%%%%%%%%%%%%%%%%%%%%%%%%%%%%%%%%%%%%%%%%%%%%%%%%
    %%%%%%%%%%%%%%%%%%%%%%%%%%%%%%%%%%%%%%%%%%%%%%%%%%%%%%%%%%%%%%%

\begin{abstract}
% Physical F/T measurement
Force and torque (F/T) sensors enable contact-aware control by providing reactive feedback, but they are often fragile and expensive.
% Sensorless methods & limitations
To overcome these limitations, sensorless methods estimate F/T or wrench solely from robot proprioception, and have shown success in slow interaction tasks such as grasping.
% Limitations in HF
However, their low-pass characteristics limit the estimation of high-frequency signals, which are critical in rapid-contact tasks such as grinding.
% Limitations in delays
Communication delays can also make their estimates outdated during deployment, but few methods address this directly.
% Propose method
To bridge these gaps, we propose a Frequency-aware Decomposition Network (FDN) to estimate vibration-rich wrench in a sensorless, multi-step-ahead manner.
% How it works?
Considering higher-frequency stochasticity, FDN spectrally decomposes the wrench horizon into a low-frequency trend and a high-frequency residual, and estimates each by pointwise regression and a learned conditional distribution, respectively.
% Additional architectural
The frequency-aware layers impose band decomposition priors on the outputs and adaptively enhance frequency amplitudes of the inputs.
% What is good?
FDN requires neither an identified robot model nor an F/T sensor during estimation.
 % How good is it? (accuracy)
On real-world grinding data from our 6-DoF hydraulic manipulator, FDN reduces high-frequency amplitude error by up to 47\% over the baselines under assumed time delays and maintains competitive low-frequency pointwise accuracy, while the baselines fail to balance these two.
% How good is it? (multi-step estimation feasibility, inference time)
We also find multi-step-ahead estimation feasible, with FDN estimating a 1,000~ms horizon within 11~ms on a single CPU thread.
% What are the rest of the paper?
Ablation studies further support our design choices. In an exploratory study, transferring wrench dynamics learned from an open-source everyday manipulation dataset reduces low-frequency error by 8\%, while high-frequency dynamics appear domain-specific.
    \end{abstract}

    %% Keywords
    \begin{keyword}
    Sensorless force estimation \sep
    Wrench estimation \sep
    Robotic grinding \sep
    Multi-step-ahead forecasting \sep
    Probabilistic forecasting \sep
    Frequency-domain learning
    \end{keyword}

\end{frontmatter}

%% file: src/body_main.tex
\input{src/sections/introduction.tex}

\input{src/sections/related_work.tex}

\input{src/sections/prelim.tex}
\input{src/sections/fdn.tex}
\input{src/sections/experiments.tex}
\input{src/sections/results.tex}
\input{src/sections/conclusions.tex}

%% file: src/sections/introduction.tex
\begin{figure*}[!h]
    \centering
    \includegraphics[width=\textwidth]{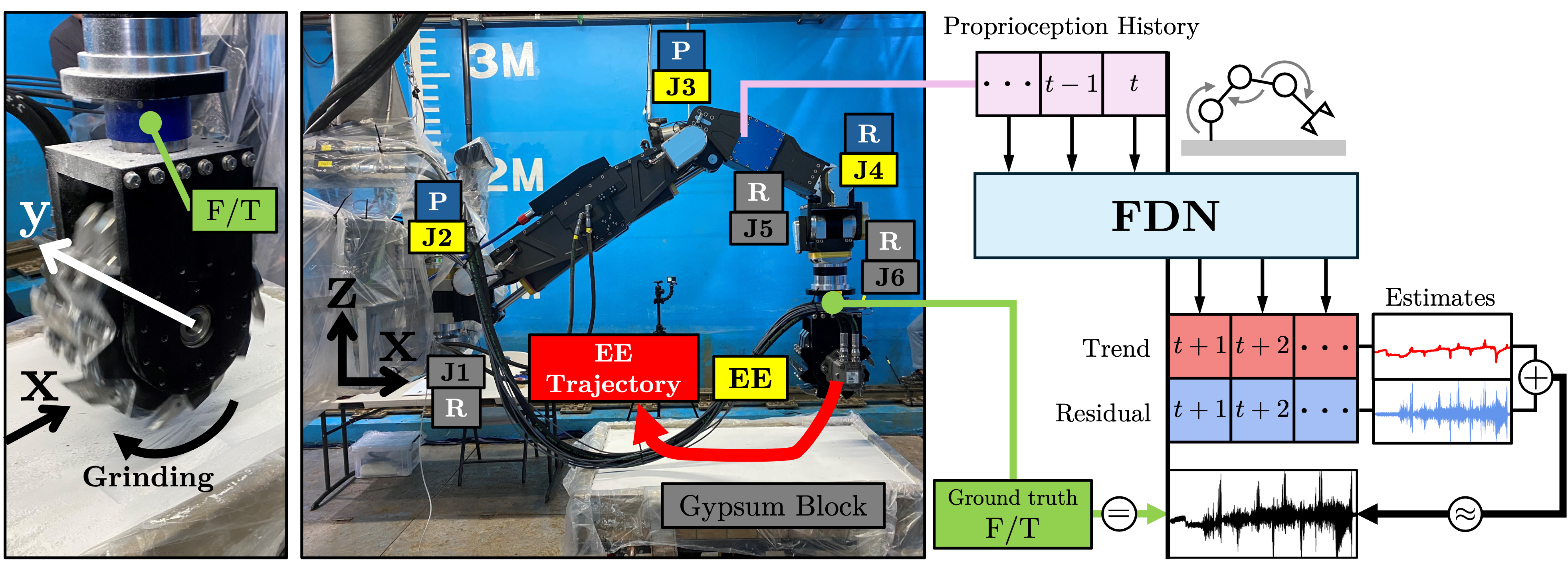}
    \caption{Illustration of our grinding task and the Frequency-aware Decomposition Network (FDN). J1 to J6 represent the joint numbers, and `EE' denotes the end-effector. Cylinder drive types are either revolute (`R') or prismatic (`P'), indicated above the joint numbers. Fixed and controlled components are colored in grey and yellow, respectively. The robot grinds the block by moving in $\mathbf{-x}$ and $\mathbf{-z}$ directions. FDN produces decomposed wrench estimates over a prediction horizon from a proprioception history, trained using ground truth from a wrist F/T sensor.
    }
    \label{fig:robot_exp}
\end{figure*}

% Overall positioning in the literature, keep motivation-focused
\section{Introduction}
\label{sec:intro}

% Importance of F/T sensing
In robotics, force and torque (F/T) sensors directly provide reactive feedback during robot-environment interaction, and have been actively employed to enable delicate, contact-aware manipulation in industrial machining \citep{norberto2002force,ji2019industrial}, robotic assembly \citep{lee2026data}, robot-assisted surgery \citep{muscolo2023force}, and embodied AI \citep{stepputtis2022system,fang2024rh20t}, as well as to provide haptic feedback for human teleoperators in surgical \citep{tholey2005force} and industrial \citep{park2025development} teleoperation systems, and to support balance control in humanoid robots \citep{kim2019multi}. 

% Hardware bottleneck of F/T sensors
Despite these benefits, F/T sensors face practical constraints at the hardware level. The high cost of commercial F/T sensors has led to the development of diverse low-cost alternatives \citep{ubeda2018design,choi2025coinft}, but practitioners usually have limited access to such non-commercial sensors. The size of the sensor is another constraint, as it must be compatible with the robot and task at hand \citep{bicchi1992criterion,li2023design}. It also directly trades off against measurement range and resolution \citep{cao2021sixaxis,kim2025parameter}, hence making an optimal selection of sensors more complicated. Furthermore, F/T sensors are generally fragile, which reduces their reliability in hazardous environments with mechanical vibrations \citep{sun2022design}, frequent impacts \citep{kim2024compact}, underwater conditions \citep{palli2015performance}, or temperature variations \citep{xiong2020six}.

% Sensorless methods: Model-based methods
To overcome these hardware limitations, sensorless approaches that estimate the interaction wrench without a physical sensor have been widely investigated. Model-based methods typically design force observers using the identified robot dynamics model. In earlier studies, \citet{hacksel1994estimation} and \citet{alcocer2003force} modeled the position error dynamics as a damped spring--mass system driven by the external wrench. \citet{murakami1993force} and \citet{eom1998disturbance} proposed disturbance observers that regard the external wrench as a lumped disturbance. \citet{deluca2003actuator} suggested a generalized momentum-based observer, and later applied it to sensorless collision detection \citep{deluca2005sensorless}. Furthermore, \citet{jung2006robust} and \citet{wahrburg2015cartesian} combined Kalman filters to account for model uncertainties and measurement noise.

% Limitations: model-based methods
Although these model-based methods are theoretically grounded and generalize well with accurately modeled dynamics, they mostly presume rigid body motions and identifiable system parameters, which do not comply with most real-world engineering scenarios. 
% Specific engineering scenarios + requirements
This makes their implementation challenging in systems including deformable contact objects \citep{yang2025force}, drive friction \citep{han2024data}, compliance \citep{armanini2023soft}, and dynamic environments \citep{huang2017vehicle}, as reliable modeling of such interaction or environment dynamics often demands identification of nonlinear effects \citep{ting2006bayesian,han2021dnn}, uncertainties \citep{han2026updating}, and system parameters \citep{wu2010overview,jang2025sid}, as well as experimental verification \citep{han2023multi}. 
Our system, shown in Fig.~\ref{fig:robot_exp}, is one such case where a hydraulic manipulator performs robotic grinding on gypsum blocks. Its dynamics accordingly involve extensive nonlinearities that are challenging to identify, such as cylinder friction dynamics with varying joint drive mechanisms, hydraulic actuation effects, and high-frequency interaction uncertainties. Building a dedicated model-based estimator is therefore highly demanding, while approximating the dynamics with simplified rigid-body components would also introduce large model mismatch, which in turn demands an iterative modeling and verification loop in practice \citep{han2024digital}.

% Sensorless method: Model-free data-driven
This reality naturally requires a model-free, data-driven approach. \citet{smith2006neural} was among the first to argue for model-free estimators and trained a multilayer perceptron (MLP) to estimate contact forces on a planar linkage arm. In more recent studies, \citet{chen2019rbf} modeled the interaction torque with a radial basis function (RBF) neural network and applied it to provide force feedback in a bilateral teleoperation system. \citet{dong2020sensorless} adopted online sparse Gaussian process regression (OSGPR) on a 3-DoF parallel robot, which updated the estimator online with a sparse set of inducing points. \citet{kruvzic2022neural} adopted a long short-term memory (LSTM) network to estimate end-effector forces on a mobile 7-DoF manipulator. \citet{pan2023adaptive} proposed an adaptive sparse general regression neural network (ASGRNN) with sparse feature selection, and \citet{pan2024graph} embedded inter-joint spatial relations in a graph robot network (GRN) to generalize across contact materials, both evaluated on 6-DoF manipulators.
% \citet{chua2021force} have also investigated vision-based force estimation for haptic feedback in robotic surgery. 
These works accurately estimated force/torque based on proprioceptive inputs, yet such estimators require training data and remain bound to the training distribution. Hybrid approaches like \citet{zahn1999learning,hu2018contact,liu2021sensorless} and \citet{hu2026physicsinformedb} retain both the generalization of model-based estimation and the flexibility of data-driven learning, but inherit the burden of both, as they still require an explicit dynamics model and its identification alongside the learned components. Therefore, the model-free data-driven method remains attractive for systems where modeling and identification are challenging.

% Gap 1: Reproducing HF fluctuations
Both families, however, face structural limitations in rapid-interaction tasks such as robotic grinding, milling, and polishing, which are prevalent in industrial and surgical robotics. Although the vibrations generated in these tasks deliver critical information about machining stability, surface quality, and contact state \citep{olsson2010cost,xu2025grinding,hu2025accurate,mu2025novel}, most sensorless estimators have limited capability to reproduce such amplitude fluctuations. 
Model-based observers inherently behave as low-pass filters, where their estimates are a smoothed and lagged version of the actual force/torque signals \citep{deluca2005sensorless,wahrburg2015cartesian,hu2026physicsinformedb}. Their achievable bandwidth is bounded by the signal-to-noise ratio in practice, as widening the estimation bandwidth amplifies measurement noise \citep{mitsantisuk2011estimation,cao2021contact}. 
Data-driven estimators are also limited in representing vibratory signals, as neural networks prioritize low-frequency components during optimization \citep{rahaman2019spectral,xu2025overview}, and models trained on such signals do not generalize well to unseen inputs \citep{lee2024cnn}. 

% Some related studies on Gap 1
Related studies by \citet{dong2020contact} and \citet{xu2025grinding} have worked on similar issues, where they adopted model-based observers to estimate contact forces in robotic polishing and grinding. However, they only retrieved the low-frequency mean of the measured force signals. Previous works in manufacturing have also mapped computer numerical control (CNC) inherent signals to high-frequency cutting forces \citep{cheng2021mechanism,li2023noninvasive,ni2024unsupervised}, where a single-axis feed drive often generates periodic cutting forces \citep{zhu2025physicsguided}. These findings, however, do not directly transfer to multi-joint robots, whose coupled joint dynamics produce more irregular contact transients.

% Gap 2: Instantaneous formulation
In addition, when the estimator is deployed in a closed loop, an estimate produced at the current instant may become outdated as it propagates through the system. Yet, existing force estimator formulations mainly target the wrench at the current instant, whereas predicting a horizon of future values from history is the standard setting in time-series forecasting \citep{de200625}. \citet{kruvzic2022neural} and \citet{pan2024graph} used the history of proprioceptive states as their inputs, but estimated forces at the current instant. \citet{sun2019bilateral} and \citet{zhang2021polishing} modeled the sequential structure of the force estimates, but did not predict them ahead of time. \citet{narayan2020developing} and \citet{scibilia2024nonlinear} formulated multi-step-ahead and one-step-ahead force estimators, but both take past force measurements as input and thus require a real sensor. \citet{azimifar2017improving} and \citet{chen2019rbf} explicitly accounted for communication delays in teleoperation, but their focus was on system-level delay awareness and robustness rather than projecting the estimate forward. Therefore, sensorless multi-step-ahead wrench estimation remains a gap in the literature, and whether it can be achieved without compromising accuracy is also uncertain.

% Design motivation
To address these gaps, we propose a Frequency-aware Decomposition Network (FDN). We hypothesize that, in our robotic grinding, the low-frequency part of the contact wrench is a deterministic trend that depends on coarse contact conditions such as the end-effector trajectory or block stiffness, while the dominant high-frequency counterpart is a stochastic variation arising from mechanical vibrations and abrasive particle interactions. Based on this insight, we spectrally decompose the wrench into a low-frequency trend and a high-frequency residual, and model them asymmetrically with deterministic pointwise regression and a probabilistic conditional distribution, respectively. The decomposition also provides a frequency cutoff prior that we impose on the estimated outputs, while the network adaptively enhances the informative frequency components of the input proprioception history. The resulting trend and residual estimates are summed at each time step to produce the final estimate, which remains valid for a time delay up to the prediction horizon. The key novelty of the proposed estimator is that, as illustrated in Fig.~\ref{fig:robot_exp}, it estimates the vibratory wrench in a multi-step-ahead manner without requiring an explicit robot dynamics model, and operates without an F/T sensor once trained. To mitigate the high cost of data acquisition in our setting, we further explore transferring learned proprioception-to-wrench dynamics from RH20T \citep{fang2024rh20t}, an open-source everyday manipulation dataset, to our hydraulic grinding setting. The main contributions of this paper are summarized as follows:

\begin{enumerate}\label{key_contributions}
    % Decomp-based modeling + FA
    \item We introduce a novel asymmetric spectral decomposition framework that separates the interaction wrench into a low-frequency trend and a high-frequency residual over the prediction horizon, and models them by pointwise regression and a conditional distribution, respectively, which enables effective representation of the high-frequency transients in the grinding wrench. Frequency-aware layers additionally impose the decomposition cutoff as a band prior on the output estimates and adaptively modulate the frequency amplitudes of the input proprioception history.
    % Multi-step-ahead formulation & validation
    \item We formulate sensorless wrench estimation as a multi-step-ahead problem over a temporal horizon so that the produced estimate remains valid after a time delay. We empirically show that this formulation does not degrade accuracy relative to instantaneous estimation, while the proposed decomposition-based modeling contributes more significantly by improving the wrench representation.
    % Real-world experiment, comparison, ablation, transfer
    \item We evaluate FDN on a real-world 6-DoF hydraulic grinding dataset using band-specific metrics under assumed time delays. We observe that FDN outperforms the baselines in the high-frequency band by up to 47\%, mitigates the accuracy trade-off between frequency bands while the others fail, and estimates over a 1,000~ms horizon within 11~ms on a single CPU thread. We further conduct an exploratory study on transferring learned wrench dynamics from RH20T \citep{fang2024rh20t} to our setting, which allowed an additional 8\% improvement in the low-frequency band. Our code and data are available at \github~and on Zenodo~\citep{lee_2026_23026021}.
\end{enumerate}

%% file: src/sections/related_work.tex
% Introduce techniques and details that I used.

\section{Related work}
\label{sec:related_works}
In this section, we provide a brief overview of the deep time-series forecasting and frequency-domain learning methods that FDN builds on, together with the works that motivate our experimental study on transferring the learned wrench dynamics across different robot datasets.

\subsection{Deep time-series forecasting}
Time-series forecasting models predict future values of a sequence based on its history and have been widely adopted for weather, traffic, and healthcare applications, as well as for commercial decision-making \citep{casolaro2023deep}. While statistical methods such as exponential smoothing and autoregressive integrated moving average (ARIMA) were long the standard in the literature \citep{de200625}, recent studies incorporate deep sequence architectures to better model nonlinear, multivariate temporal dynamics \citep{li2024deep}. In earlier works, recurrent and convolutional networks were frequently employed, but they were structurally limited in extracting global temporal features from a long sequence due to unstable gradients and fixed local receptive fields, respectively \citep{casolaro2023deep}. 

The subsequent emergence of the attention mechanism \citep{bahdanau2014neural} and the Transformer \citep{vaswani2017attention} architecture mitigated these issues and led to the development of diverse Transformer-based forecasting models. Informer \citep{zhou2021informer} focused on an efficient encoder--decoder mechanism, where its generative-style decoder produced a sequence output directly without autoregressive decoding, and sparse self-attention encoded long time-series inputs with a significantly lower computational budget. Autoformer \citep{wu2021autoformer} explicitly integrated moving-average-based trend--seasonal decomposition and an autocorrelation mechanism to capture long-term periodicity. FEDformer \citep{zhou2022fedformer} extended the Autoformer's moving-average decomposition with a mixture-of-experts (MoE) \citep{jacobs1991adaptive} mechanism to adaptively combine multiple moving-average kernels, while adopting frequency-based blocks to extract spectral features from sequences. PatchTST \citep{nie2022time} introduced a channel-independent patching mechanism, which efficiently retained local semantic information at the segment level for both long-term forecasting and representation learning. Furthermore, iTransformer \citep{liu2023itransformer} simply switched the self-attention dimension from the time axis to the channel axis to capture cross-variable dependencies, and outperformed state-of-the-art models with only the vanilla Transformer.

Beyond the backbone architecture, another line of work concerns how forecasting models handle their inputs and outputs. \citet{kim2021reversible} addressed distribution shift in time-series data by adding a reversible sample-level normalization technique, which normalizes the input and inverts the statistics at the output. Probabilistic forecasting models focus on representing uncertainties in their predictions. \citet{salinas2020deepar} trained RNNs to output conditional distribution parameters of the output variables, and \citet{lim2021temporal} designed a Transformer-based model to predict quantile values over the horizon. Integrating exogenous variables, i.e., external variables that influence the output variables, has also received increasing attention \citep{lim2021temporal,tayal2024exotst,wang2024timexer}. However, this line of work regards them as auxiliary information to their primary input, namely the history of the output variables, and does not support the fully exogenous input required by the sensorless setting. We therefore introduce a channel-mixing projection and a modified reversible normalization in Section~\ref{subsec:input_repr} to adapt a time-series forecasting backbone to the fully exogenous setting.

\subsection{Frequency-domain learning}\label{subsec:freqlearning}

Rapid-contact tasks generate high-frequency vibrations that carry useful task-related information, as previously discussed in Section~\ref{sec:intro}. However, learning them is often impeded by a characteristic of neural networks known as spectral bias, which refers to their strong preference for low-frequency components of a target function during optimization \citep{rahaman2019spectral}. In computer vision and nonlinear dynamics, \citet{tancik2020fourier} and \citet{lee2024cnn} showed that this characteristic can be addressed by mapping the inputs to higher-frequency features. A further step beyond the feature space mapping leads to operating in the frequency domain itself, which time-series forecasting has increasingly adopted to enhance spectral features, increase computational efficiency, and improve high-frequency learning \citep{yi2025surveya,zhang2025time}. Specifically, FEDformer \citep{zhou2022fedformer} developed this direction in time-series forecasting by applying random mode selection and a frequency-domain attention mechanism to its progressively decomposed feature sequences. Fredformer \citep{piao2024fredformer} addressed the spectral bias more directly by balancing disproportionately large low-frequency energy with the high-frequency components through local spectrum normalization. FreqMoE \citep{liu2025freqmoe} also targeted a similar problem and applied MoE over local frequency bands to prevent the model from neglecting critical but low-energy high-frequency patterns.
These frequency-based forecasting models commonly apply band-specialized processing to their input features and effectively capture high-frequency patterns from them, but eventually aggregate all bands to produce a single final pointwise forecast and overlook output stochasticity in higher-frequency bands. By contrast, FDN separately models low- and high-frequency band outputs reflecting their distinct statistical properties, and applies frequency-domain refinement to both its input and output, as detailed in Sections~\ref{subsec:probheads} and~\ref{subsubsec:frequency_awareness}.

\subsection{Transfer of learned dynamics across physical systems}

Data-driven models learn the dynamics of a physical system directly from data, fitting the parameters and configurations of the target system. However, our prior knowledge of physics gives us general laws that can be applied across different systems, and this naturally raises a question of whether the learned dynamics can be reused outside the trained configurations. \citet{yilmaz2022transfer} has pursued a similar direction in robotics, using an LSTM model as a joint torque estimator for da Vinci surgical robots. They pretrained the model on free-space motion datasets available from the community and successfully transferred the learned torque dynamics to different da Vinci robots through short adaptations. In fluid mechanics, \citet{mccabe2024multiple} pretrained a Transformer-based video prediction model on a large-scale fluid simulation dataset, and demonstrated that the pretrained dynamics can generalize to unseen boundary conditions, equations, and parameters, and even outperform task-specific baselines. Relevant approaches also appear in molecular \citep{batatia2025foundation} and structural dynamics \citep{yang2025few}, where the learned force dynamics transfer across different materials and mechanical systems, respectively.

In a broader context, transfer is not only limited to learned dynamics but also extends to forecasting representations in time-series modeling \citep{miller2024survey} and control policies in robot learning \citep{jaquier2025transfer}. Particularly in robotics, researchers are actively collaborating to build an open-source ecosystem following the success of large-scale pretraining in vision-language-action models \citep{zitkovich2023rt,kim2024openvla}. This has stimulated the extensive release of public robot datasets, like RH20T \citep{fang2024rh20t} and REASSEMBLE \citep{sliwowski2025reassemble}, which even contain force/torque measurements along with standard proprioception and vision data. The availability of such datasets, together with the directions above, naturally motivates learning and transferring wrench dynamics from the open-source robot datasets, which we explore in Section~\ref{subsec:transfer}.

%% file: src/sections/prelim.tex
\section{Preliminaries}
\label{sec:prelim}
To formulate the multi-step-ahead wrench estimation problem, we start from the instantaneous rigid-body dynamics of a robot and extend the dynamics over a temporal horizon with a numerical integration of a state-space representation.

\subsection{Robot dynamics and wrench estimation}
% Dynamic model
The dynamic model of a robot manipulator is generally expressed as:
\begin{equation}
    \label{eq:dynamics}
    \mathrm{M}(\boldsymbol{q})\boldsymbol{\ddot{q}}+\mathrm{C}(\boldsymbol{q},\boldsymbol{\dot{q}})\boldsymbol{\dot{q}}+\mathrm{G}(\boldsymbol{q})+\mathrm{F}(\boldsymbol{\dot{q}})=\boldsymbol{\tau}+\mathrm{J}(\boldsymbol{q})^T \boldsymbol{W}
\end{equation}
where $\boldsymbol{q},\boldsymbol{\dot{q}}$, and $\boldsymbol{\ddot{q}}$ represent joint position, velocity, and acceleration vectors, respectively. $\mathrm{M}(\boldsymbol{q})$ denotes a positive definite inertia matrix, $\mathrm{C}(\boldsymbol{q},\boldsymbol{\dot{q}})$ is the Coriolis and centrifugal matrix, $\mathrm{G}(\boldsymbol{q}), \mathrm{F}(\boldsymbol{\dot{q}})$ denote gravity and friction terms. $\boldsymbol{\tau}$ is joint actuator torque, $\mathrm{J}(\boldsymbol{q})$ is the robot Jacobian, and $\boldsymbol{W}=[\boldsymbol{F};\boldsymbol{M}]$ is Cartesian space interaction wrench.

% Model-based estimation and data-driven method
The inverse dynamics model of Eq.~\ref{eq:dynamics} directly gives the wrench or force estimation:
\begin{equation}
    \label{eq:inv_dynamics}
    \boldsymbol{W} = \mathrm{J}^{-T}(\boldsymbol{q})   \big(   \mathrm{M}(\boldsymbol{q})\boldsymbol{\ddot{q}}+\mathrm{C}(\boldsymbol{q},\boldsymbol{\dot{q}})\boldsymbol{\dot{q}}+\mathrm{G}(\boldsymbol{q})+\mathrm{F}(\boldsymbol{\dot{q}})-\boldsymbol{\tau}    \big)
\end{equation}
where the joint acceleration $\ddot{\boldsymbol{q}}$ is usually not measurable and difficult to estimate due to noise. $\mathrm{J}(\boldsymbol{q})$ is assumed nonsingular. Observer-based methods reviewed in Section~\ref{sec:intro} avoid this direct inversion by treating the external wrench as a disturbance to be reconstructed, yet they require the terms of Eq.~\ref{eq:dynamics} to be modeled and identified beforehand. Nevertheless, Eq.~\ref{eq:inv_dynamics} specifies which variables determine the wrench, namely the joint states and the actuation.

\subsection{Multi-step-ahead wrench estimation}
\label{subsec:model}
We extend the instantaneous dynamics in Eq.~\ref{eq:inv_dynamics} over a temporal horizon to characterize the temporal dependence of a multi-step-ahead wrench sequence $[\boldsymbol{W}_{t+1},\boldsymbol{W}_{t+2},\cdots]$. To solve the second-order dynamics, we define a state vector $\boldsymbol{y}=[\boldsymbol{q},\boldsymbol{\dot{q}}]^T$ and rewrite Eq.~\ref{eq:dynamics} as a first-order system:
\begin{equation}
    \label{eq:statespace}
    \boldsymbol{\dot{y}}
    = h(\boldsymbol{y},\boldsymbol{\tau},\boldsymbol{W})
\end{equation}
where $\boldsymbol{\dot{y}}=[\boldsymbol{\dot{q}}, \boldsymbol{\ddot{q}}]^T$ and
\begin{equation}\label{eq:acceleration}
    \begin{aligned}
        \boldsymbol{\ddot{q}} & = \mathrm{M}^{-1}(\boldsymbol{q})   \left(  \boldsymbol{\tau}+\mathrm{J}(\boldsymbol{q})^T \boldsymbol{W}-\mathrm{C}(\boldsymbol{q},\boldsymbol{\dot{q}})\boldsymbol{\dot{q}}-\mathrm{G}(\boldsymbol{q})-\mathrm{F}(\boldsymbol{\dot{q}}) \right) \\
    \end{aligned}
\end{equation}
Then, the solution of Eq.~\ref{eq:statespace} can be obtained via recursive numerical integration:
\begin{equation}
    \label{eq:robot_transition}
    \boldsymbol{y}_{t+1}= \Phi_{\Delta t} (\boldsymbol{y}_t, \boldsymbol{\dot{y}}_t)
\end{equation}
where $\Phi_{\Delta t}$ denotes a numerical integrator with a time step size $\Delta t$. Let $\psi$ denote the inverse dynamics mapping defined by Eq.~\ref{eq:inv_dynamics} and $\boldsymbol{W} =\psi(\boldsymbol{y},\boldsymbol{\dot{y}},\boldsymbol{\tau})$. By combining Eq.~\ref{eq:statespace} and \ref{eq:robot_transition} into $\psi$ at time $t+1$, we obtain the one-step dependence:

\begin{equation}
    \label{eq:wrench_transition}
    \begin{aligned}
        \boldsymbol{W}_{t+1} & = \psi  \big(   \Phi_{\Delta t}  \big(  \boldsymbol{y}_t, h(\boldsymbol{y}_t,\boldsymbol{\tau}_t,\boldsymbol{W}_t) \big) ,\boldsymbol{\dot{y}}_{t+1},\boldsymbol{\tau}_{t+1}  \big)
    \end{aligned}
\end{equation}
which implies temporal dependency between one-step-ahead wrench $\boldsymbol{W}_{t+1}$ and current joint states $\boldsymbol{y}_t=[\boldsymbol{q}_t, \boldsymbol{\dot{q}}_t]^T$, actuator torque $\boldsymbol{\tau}_t$, and wrench $\boldsymbol{W}_t$. For a sufficiently small $\Delta t$ and locally smooth states, we can adopt local approximations $\boldsymbol{\dot{y}}_{t+1}\approx \boldsymbol{\dot{y}}_{t}$ and $\boldsymbol{\tau}_{t+1}\approx \boldsymbol{\tau}_{t}$ and rewrite Eq.~\ref{eq:wrench_transition} as:
\begin{equation}
    \label{eq:temporal_dependency}
    \boldsymbol{W}_{t+1} = \mathcal{T}_{\Delta t}(\boldsymbol{q}_t, \boldsymbol{\dot{q}}_t,\boldsymbol{\ddot{q}}_t, \boldsymbol{\tau}_t,\boldsymbol{W}_t) + \boldsymbol{e}_{t+1}
\end{equation}
where we define an approximate one-step transition $\mathcal{T}_{\Delta t}$ and an approximation error term $\boldsymbol{e}_{t+1}$. Letting $\mathcal{X}_t=[\boldsymbol{q}_t, \boldsymbol{\dot{q}}_t,\boldsymbol{\ddot{q}}_t, \boldsymbol{\tau}_t]$ and rolling out $\mathcal{T}_{\Delta t}$ yields:
\begin{equation}
    \label{eq:rollout}
    \boldsymbol{W}_{t+j} \approx \mathcal{T}_{\Delta t}(\mathcal{X}_{t+j-1}, \boldsymbol{W}_{t+j-1}), \quad j=1, \cdots, T
\end{equation}
which suggests that the wrench over a $T$-step-ahead horizon $\boldsymbol{W}_{t+1:t+T}=[\boldsymbol{W}_{t+1},\cdots,\boldsymbol{W}_{t+T}]$ approximately depends on the trajectories of recursively propagated motion, actuator torque, and wrench up to time $t+T-1$.

% Practical limitation
This approximated recursive dependency formulates multi-step-ahead estimation, but the rollout is practically infeasible due to partial observability. Specifically, we presume that (i) the wrench is not measurable and (ii) the states are observed up to the present time $t$ only. Under this observation, we can alternatively model a conditional distribution of the wrench over a $T$-step horizon given a finite $L$-step history of observable states $\boldsymbol{x}_t = [\boldsymbol{q}_t, \boldsymbol{\dot{q}}_t,\boldsymbol{\ddot{q}}_t, \boldsymbol{u}_t]$:
\begin{equation}
    \label{eq:forecasting_distribution}
    \boldsymbol{W}_{t+1:t_f} \sim \mathcal{P}(\cdot\mid \boldsymbol{x}_{t_h:t})
\end{equation}
where $t_h=t-L+1$ is the history start index and $t_f=t+T$ is the final estimation index. The observable input in Eq.~\ref{eq:forecasting_distribution} differs from the input variables required in Eq.~\ref{eq:rollout} in three respects: (i) $\boldsymbol{\tau}$ is replaced with measurable actuation signals $\boldsymbol{u}$ such that $\boldsymbol{\tau}=\boldsymbol{\tau}(\boldsymbol{u})$, where $\boldsymbol{u}$ can be motor torque or current in motor-actuated robots, or joint differential pressure $\Delta\boldsymbol{p}$ in hydraulic manipulators, (ii) unmeasurable $\boldsymbol{W}$ is removed from the input, and (iii) time steps are observable up to $t$ only. To address this discrepancy, we assume that (i) $\boldsymbol{u}$ serves as a measurable proxy for $\boldsymbol{\tau}$, (ii) the internal states retain information about recent wrench effects, motivated by Eq.~\ref{eq:dynamics}, and (iii) we focus on a short-term horizon $T$ where the finite state history $\boldsymbol{x}_{t_h:t}$ can serve as a valid summary of recent interaction trends. Accordingly, the formulation targets short-horizon estimation within a locally consistent operating conditions, rather than anticipating abrupt future commands or environmental changes that are not inferable from the observed history. We then approximate $\mathcal{P}$ with a model $\mathcal{M}_\theta$:
\begin{equation}\label{eq:model}
    \hat{\boldsymbol{W}}_{t+1:t_f}\sim \mathcal{M}_\theta(\cdot\mid\boldsymbol{x}_{t_h:t})
\end{equation}
where $\theta$ denotes the parameters of the model. 
$\mathcal{M}_\theta$ thus estimates the wrench over the next $T$ steps from the measurable signals alone, without the system matrices in Eq.~\ref{eq:inv_dynamics} or the unmeasurable wrench in Eq.~\ref{eq:rollout}.

% Predictable regime
Eq.~\ref{eq:model} does not specify how $\mathcal {M}_\theta$ should generate an estimation over a horizon, although the equation is derived from the recursive relation in Eq.~\ref{eq:rollout}. $\mathcal{M}_\theta$ could either be recursive or non-recursive, depending on its architecture. The formulation also does not determine how the wrench should be represented over the horizon. We address both of these in the following section.

%% file: src/sections/fdn.tex
\section{Frequency-aware Decomposition Network}\label{sec:fpn}

\begin{figure*}[!t]
    \includegraphics[width=\textwidth]{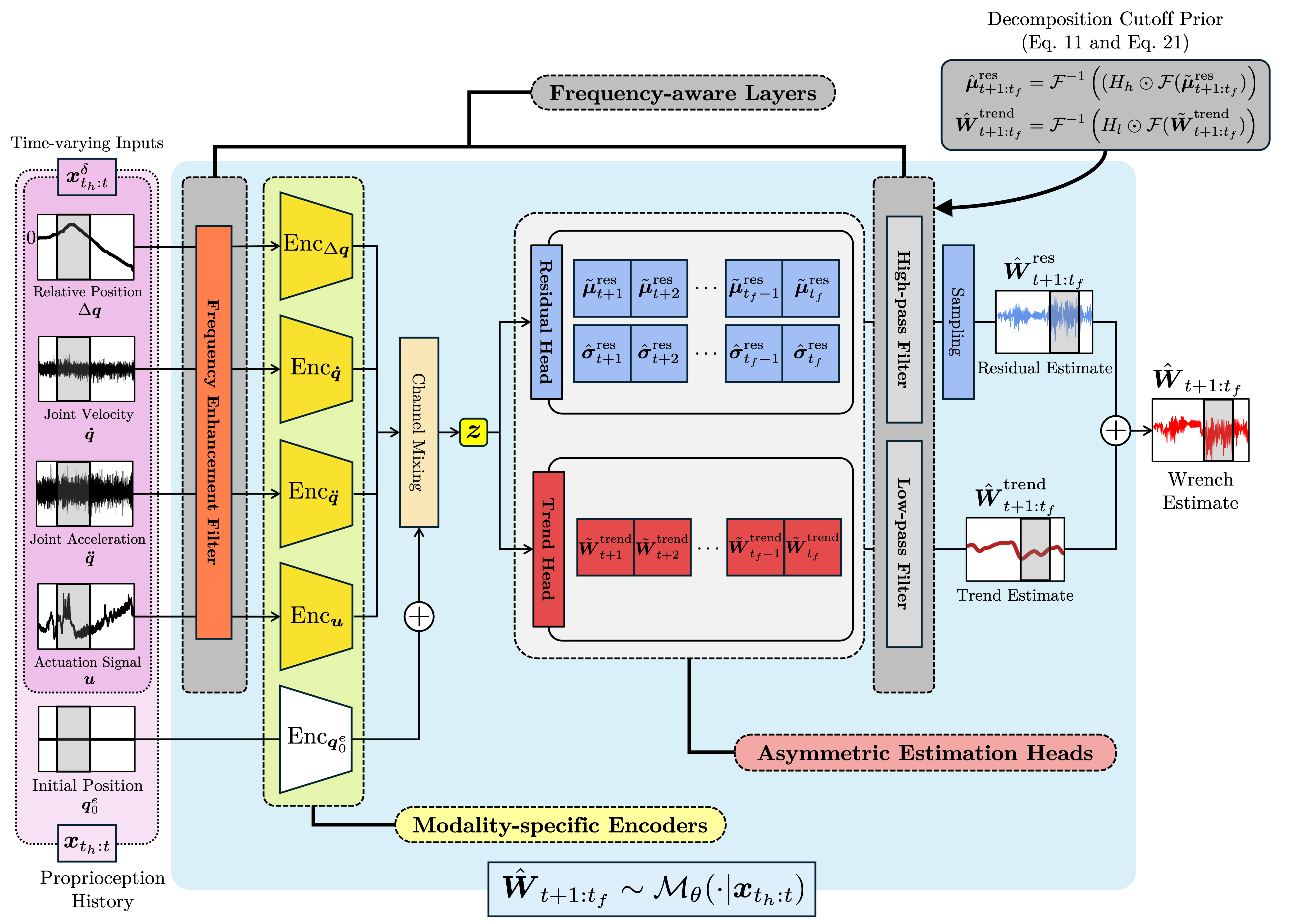}
    \caption{Illustration of the proposed Frequency-aware Decomposition Network (FDN).}
    \label{fig:model}
\end{figure*}

% Keep focus still on Eq.10
We propose the Frequency-aware Decomposition Network as a data-driven solution to $\mathcal{M}_\theta$ in Eq.~\ref{eq:model}. % Addressing variable issues in Eq 10
% Figure 2
Fig.~\ref{fig:model} illustrates the proposed FDN. Following the distinct low- and high-frequency characteristics of the grinding wrench discussed in Section~\ref{sec:intro}, we spectrally decompose the wrench over each horizon and represent the two components with pointwise regression and a conditional distribution, respectively, instead of applying a single representation form over the full frequency band. FDN produces both wrench components from $\boldsymbol{x}_{t_h:t}$ in a single forward pass without recursion, and also employs frequency-aware layers motivated by the frequency-domain learning methods reviewed in Section~\ref{sec:related_works}.

For the actuation signal $\boldsymbol{u}$ in the input $\boldsymbol{x}_{t_h:t}$, it is known that its mapping to  $\boldsymbol{\tau}(\boldsymbol{u})$ in hydraulic manipulators depends on cylinder characteristics such as drive geometry and friction \citep{han2024data}. Nevertheless, we do not explicitly identify such system-specific relations in $\mathcal{M}_\theta$, and let it capture the actuation-signal-to-torque mapping fully from measurable data. Also, position derivatives $\boldsymbol{\dot{q}}$ and $\boldsymbol{\ddot{q}}$ in our work are obtained numerically from the measured joint position $\boldsymbol{q}$, as in previous data-driven estimators \citep{smith2006neural,dong2020contact}. 
Note that unlike the algebraic inversion in Eq.~\ref{eq:inv_dynamics}, where the derivative errors are scaled directly by the system matrices, the derivatives in Eq.~\ref{eq:model} enter $\mathcal{M}_\theta$ as normalized input channels whose contribution is learned during training.
The effect of using these numerical derivatives is examined in Section~\ref{subsec:input_ablation}.

% Overview
\subsection{Horizon-level spectral decomposition}
\label{subsec:decomp}
% Ground truth decomposition
% To apply asymmetric modeling to different wrench components
To model lower- and higher-frequency bands separately, we first spectrally decompose $\boldsymbol{W}_{t+1:t_f}$ into trend $\boldsymbol{W}^{\mathrm{trend}}_{t+1:t_f}$ and residual $\boldsymbol{W}^{\mathrm{res}}_{t+1:t_f}$ over each $T$-step horizon, and then let our model $\mathcal{M}_\theta$ yield their estimates. We presume that $\boldsymbol{W}$ is denoised at the episode level with a cutoff frequency $f_c^{\mathrm{dn}}$ prior to the decomposition as a preprocessing step. For each estimation time $t$, the ground-truth trend and residual sequences are defined using horizon-level spectral decomposition as:
\begin{equation}
    \begin{aligned}
        \boldsymbol{W}^{\mathrm{trend}}_{t+1:t_f} & = \mathrm{FPF}_{\mathrm{low}} ( \boldsymbol{W}_{t+1:t_f} )             \\
        \boldsymbol{W}^{\mathrm{res}}_{t+1:t_f}   & = \boldsymbol{W}_{t+1:t_f} - \boldsymbol{W}_{t+1:t_f}^{\mathrm{trend}} \\
    \end{aligned}
    \label{eq:decomp}
\end{equation}
where
\begin{equation}\label{eq:fpf_low}
    \mathrm{FPF}_{\mathrm{low}} (\cdot) = \mathcal{F}^{-1}\left(H_l(f;f_c)\odot\mathcal{F}(\cdot)\right)
\end{equation}
is a differentiable, non-recursive low-pass filter where
\begin{equation}\label{eq:filter_response}
    H_l(f;f_c) = 1/\sqrt{1+(f/f_c)^{2r}}, \quad f\in[0, f_{\mathrm{Nyq}}]
\end{equation}
\noindent
is a real-valued Butterworth low-pass response \citep{butterworth1930theory} with cutoff frequency $f_c$. $f_{\mathrm{Nyq}}$ is the Nyquist frequency. $\mathcal{F},\mathcal{F}^{-1}$ are the Fast Fourier Transform (FFT) and its inverse, and $\odot$ is an elementwise multiplication operator. $r$ is the order of the response. In our work, we set $r=8$ and apply both-sided reflection padding before the FFT and center-crop the inverse-transformed sequence to reduce boundary artifacts.

The FDN model $\mathcal{M}_{\theta}$ then learns to estimate wrench decompositions given $\boldsymbol{x}_{t_h:t}$ as:
\begin{equation}\label{eq:decomposed_output}
    \{\hat{\boldsymbol{W}}_{t+1:t_f}^{\mathrm{trend}},\hat{\boldsymbol{W}}_{t+1:t_f}^{\mathrm{res}}\} \sim \mathcal{M}_{\theta}(\cdot\mid\boldsymbol{x}_{t_h:t})
\end{equation}
The final estimate $\hat{\boldsymbol{W}}_{t+1:t_f}$ is obtained by summing these trend and residual estimates:
\begin{equation}
    \hat{\boldsymbol{W}}_{t+1:t_f}                                                                   =\hat{\boldsymbol{W}}_{t+1:t_f}^{\mathrm{trend}}+\hat{\boldsymbol{W}}_{t+1:t_f}^{\mathrm{res}}
\end{equation}

\subsection{Modality-specific encoders}
\label{subsec:input_repr}

% Intro & redefine input vector
To reduce the model's sensitivity to initial position shifts and to separately handle the heterogeneous input modalities, we redefine the input vector $\boldsymbol{x}_t$ and encode each modality with a separate modality-specific encoder. The redefinition divides $\boldsymbol{q}_t$ into relative joint positions $\Delta \boldsymbol{q}$ and episode initial positions $\boldsymbol{q}_0^e$:

\begin{equation}\label{eq:input_rel}
    \boldsymbol{x}_t=[\Delta\boldsymbol{q}_t, \boldsymbol{\dot{q}}_t,\boldsymbol{\ddot{q}}_t, \boldsymbol{u}_t,\boldsymbol{q}_0^e]\in\mathbb{R}^{5n}
\end{equation}
for a $n$-DoF robot, where
\begin{equation}
    \label{eq:rel_kine}
    \Delta \boldsymbol{q}_t = \boldsymbol{q}_t-\boldsymbol{q}_0^{e}
\end{equation}
% Encoding mechanism
Then, given an input history $\boldsymbol{x}_{t_h:t}\in\mathbb{R}^{5n\times L}$, our model computes a sequence representation using four modality-specific PatchTST \citep{nie2022time} encoders $\mathrm{Enc}_{\Delta \boldsymbol{q}}, \mathrm{Enc}_{\boldsymbol{\dot{q}}}, \mathrm{Enc}_{\boldsymbol{\ddot{q}}}$, and $\mathrm{Enc}_{\boldsymbol{u}}$ for time-varying modalities and a MLP $\mathrm{Enc}_{\boldsymbol{q}_0^e}$ for episode-varying initial positions. 

For the time-varying subset $\boldsymbol{x}_{t_h:t}^\delta=[\Delta\boldsymbol{q}_{t_h:t}, \boldsymbol{\dot{q}}_{t_h:t},\boldsymbol{\ddot{q}}_{t_h:t}, \boldsymbol{u}_{t_h:t}]\in\mathbb{R}^{4n\times L}$, we first normalize it with temporal mean and standard deviation, enhance them in the frequency domain, which will be described in Section~\ref{subsubsec:frequency_awareness}, then patch-embed $\mathrm{FEF}(\boldsymbol{x}_{t_h:t}^{\delta})$ to a latent dimension $D$ with patch length $P$ and stride $S=P$ using modality-specific embedding layers. Subsequently, we obtain representations $\{z_{\Delta \boldsymbol{q}},     z_{ \boldsymbol{\dot{q}}},   z_{ \boldsymbol{\ddot{q}}},    z_{ \boldsymbol{u}}   \}\in\mathbb{R}^{n\times N\times D}$ from the embeddings using modality-specific Transformer \citep{vaswani2017attention} encoders, where $N=\lfloor (L-P)/S\rfloor+2$ is the number of patches, and apply the inverse normalization to the resulting representations $\{z_{\Delta \boldsymbol{q}},     z_{ \boldsymbol{\dot{q}}},   z_{ \boldsymbol{\ddot{q}}},    z_{ \boldsymbol{u}}   \}$. Here, note that this is a modified implementation of the RevIN \citep{kim2021reversible} adopted in PatchTST \citep{nie2022time}. The original RevIN assumes identical input and output channels and denormalizes the final output with input statistics. This cannot be achieved in our setting, so we instead denormalize the encoder representations, which retain the input channel dimension.

For $\boldsymbol{q}_0^e$, we compute a $D$-dimensional representation $z_{\boldsymbol{q}_0^e}$ with $\mathrm{Enc}_{\boldsymbol{q}_0^e}$, broadcast it across patches and channels, and add it to $z_{\Delta \boldsymbol{q}}$. Then, we concatenate the representations along the channel dimension, yielding $z^{\prime}=[z_{\Delta \boldsymbol{q}}+z_{\boldsymbol{q}_0^e},     z_{\boldsymbol{\dot{q}}},   z_{\boldsymbol{\ddot{q}}},    z_{\boldsymbol{u}}   ]\in\mathbb{R}^{4n\times N\times D}$. Finally, we apply a channel-mixing linear projection to $z^{\prime}$ to map the channel dimension from $4n$ to the 6 wrench channels, and obtain the head-shared representation $z\in\mathbb{R}^{6\times N\times D}$.

\subsection{Asymmetric estimation heads}
\label{subsec:probheads}
Using two separate linear heads, we estimate the trend and residual components obtained from Eq.~\ref{eq:decomp} in parallel from the head-shared representation $z$.
We first flatten the patch and latent dimensions of $z$:
\begin{equation}
    \bar{z}=\mathrm{flatten}_{N,D}(z)\in\mathbb{R}^{6\times ND}
\end{equation}
\noindent
Then, we directly map $\bar{z}$ to trend and residual sequences over the $T$-step horizon. 

Specifically, for the residual, we predict the time- and channel-independent Gaussian parameters by linearly projecting the flattened dimension $ND$ to $T$ as:
\begin{equation}\label{eq:residual_heads}
    \begin{aligned}
        \tilde{\boldsymbol{\mu}}^{\mathrm{res}}_{t+1:t_f} & =\bar{z}W_\mu+b_\mu \\
        \hat{\mathbf{v}}^{\mathrm{res}}_{t+1:t_f}         & =\bar{z}W_v+b_v
    \end{aligned}
\end{equation}
\noindent
where $W_\mu,W_v\in\mathbb{R}^{ND\times T}$ and $b_\mu,b_v\in\mathbb{R}^T$. The outputs $\tilde{\boldsymbol{\mu}}^{\mathrm{res}}_{t+1:t_f},\hat{\mathbf{v}}^{\mathrm{res}}_{t+1:t_f}\in\mathbb{R}^{6\times T}$ denote predicted mean and log variances, and $\hat{\boldsymbol{\sigma}}^{\mathrm{res}}_{t+1:t_f}=\mathrm{exp}(\hat{\mathbf{v}}^{\mathrm{res}}_{t+1:t_f}/2)$. 

Similarly, we estimate the trend by a linear projection:
\begin{equation}\label{eq:trend_head}
    \tilde{\boldsymbol{W}}^{\mathrm{trend}}_{t+1:t_f}=\bar{z}W_{\mathrm{trend}}+b_{\mathrm{trend}}
\end{equation}
\noindent
where $W_{\mathrm{trend}}\in\mathbb{R}^{ND\times T}$, $b_{\mathrm{trend}}\in\mathbb{R}^T$, and $\tilde{\boldsymbol{W}}^{\mathrm{trend}}_{t+1:t_f}\in\mathbb{R}^{6\times T}$. These asymmetric estimation heads allow FDN to represent high-frequency transients as a conditional Gaussian distribution, while maintaining standard pointwise regression for the low-frequency trends. The non-autoregressive, direct structure also facilitates efficient computation of the multi-step wrench \citep{nie2022time}.

\subsection{Frequency-aware layers}
\label{subsubsec:frequency_awareness}

\begin{figure*}[!tb]
    \includegraphics[width=\textwidth]{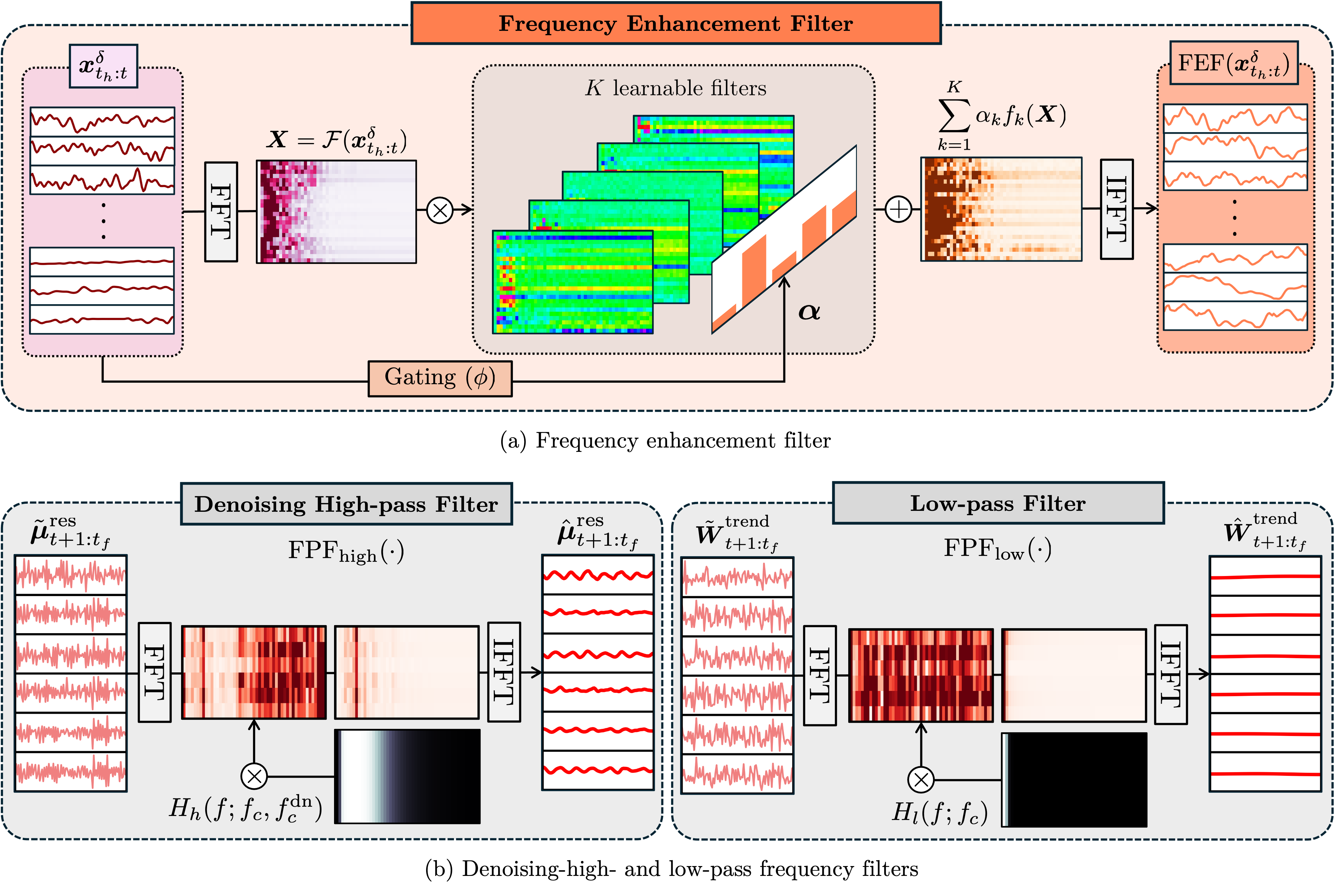}
    \caption{Frequency-aware layers. FFT and IFFT denote the Fast Fourier Transform and its inverse. (a) A learnable frequency enhancement filter ($\mathrm{FEF}$) layer.
    (b) Denoising high-pass filter ($\mathrm{FPF}_{\mathrm{high}}$) and low-pass filter ($\mathrm{FPF}_{\mathrm{low}}$).
    }
    \label{fig:fourier}
\end{figure*}

% Frequency refinement
We incorporate frequency-aware layers into FDN, as visualized in Fig.~\ref{fig:fourier}. The frequency pass filter ($\mathrm{FPF}$) imposes the cutoffs $f_c$ and $f_c^{\mathrm{dn}}$ from Eq~\ref{eq:decomp} on the estimated outputs, while the frequency enhancement filter ($\mathrm{FEF}$) adaptively enhances the frequency components of the input. Detailed mechanisms of each layer are described in the following sections.

\subsubsection{Frequency pass filters}
To constrain frequency bandwidth of each $\tilde{\boldsymbol{W}}_{t+1:t_f}^{\mathrm{trend}}$ and $\tilde{\boldsymbol{\mu}}_{t+1:t_f}^{\mathrm{res}}$ from Eqs.~\ref{eq:trend_head}~and~\ref{eq:residual_heads} as in their ground-truths in $[0,f_c]$ and $[f_c,f_c^{\mathrm{dn}}]$, respectively, we filter the estimates as:
\begin{equation}\label{eq:freqpassfilt}
    \begin{aligned}
        \hat{\boldsymbol{W}}^{\mathrm{trend}}_{t+1:t_f} & = \mathrm{FPF}_{\mathrm{low}} (\tilde{\boldsymbol{W}}^{\mathrm{trend}}_{t+1:t_f})  \\
        \hat{\boldsymbol{\mu}}^{\mathrm{res}}_{t+1:t_f} & = \mathrm{FPF}_{\mathrm{high}} (\tilde{\boldsymbol{\mu}}^{\mathrm{res}}_{t+1:t_f})
    \end{aligned}
\end{equation}
where
\begin{equation}
    \begin{aligned}
        \mathrm{FPF}_{\mathrm{high}} (\cdot) & = \mathcal{F}^{-1}\left(H_h(f;f_c,f_c^{\mathrm{dn}})\odot\mathcal{F}(\cdot)\right)
    \end{aligned}
\end{equation}
is a denoising high-pass filter and
\begin{equation}
    H_h(f;f_c,f_c^{\mathrm{dn}})=\big(\mathbf{1}-H_l(f;f_c) \big) H_l(f;f_c^{\mathrm{dn}})
\end{equation}
\noindent
is a band-pass response between $f_c$ and $f_c^{\mathrm{dn}}$ ($f_c\leq f_c^{\mathrm{dn}}$). After filtering the predictive mean, we finally sample the residual at each step independently across time as:
\begin{equation}\label{eq:res_sampling}
    \hat{\boldsymbol{W}}^{\mathrm{res}}_{t+j} = \hat{\boldsymbol{\mu}}^{\mathrm{res}}_{t+j} + \epsilon_{j}\odot\mathrm{exp}\left( \frac{\hat{\mathbf{v}}^{\mathrm{res}}_{t+j}}{2} \right)  \quad \forall{j=1,\cdots,T}\\
\end{equation}
\noindent
to generate a residual sequence $\hat{\boldsymbol{W}}^{\mathrm{res}}_{t+1:t_f}\in\mathbb{R}^{6\times T}$. Here, $\epsilon_{j} \sim \mathcal{N}(\mathbf{0},I)$ is a white noise vector in $\mathbb{R}^6$.

We use the filtered estimates in Eq.~\ref{eq:freqpassfilt} to compute the training loss in Eq.~\ref{eq:loss_total} since the filtering operations are differentiable. Here, $\hat{\boldsymbol{W}}^{\mathrm{res}}_{t+1:t_f}$ can be additionally filtered using $\mathrm{FPF}_{\mathrm{high}}$ for sample-level refinement, but we omit this design as it would require additional filtering during evaluation and complicate metric computation in Section~\ref{subsec:eval_method}.

\subsubsection{Frequency enhancement filter}

Motivated by the frequency-domain time-series models and the mixture-of-experts (MoE) mechanism in Section~\ref{sec:related_works}, we apply a frequency enhancement filter to the time-varying input $\boldsymbol{x}_{t_h:t}^\delta=[\Delta\boldsymbol{q}_{t_h:t}, \boldsymbol{\dot{q}}_{t_h:t},\boldsymbol{\ddot{q}}_{t_h:t}, \boldsymbol{u}_{t_h:t}]$. This constructs multiple candidate spectra from $\mathcal{F}(\boldsymbol{x}_{t_h:t}^\delta)$ via learnable frequency filters, combines them using input-conditioned gating weights, and reverts the combination to the time domain.

% single filter definition & phase preservation
We first define a single learnable frequency response:
\begin{equation}
    \label{eq:complexfilt}
    f(\boldsymbol{X})=\boldsymbol{X}\odot \mathrm{softplus}(W)
\end{equation}

where $\boldsymbol{X}=\mathcal{F}(\boldsymbol{x}_{t_h:t}^\delta)\in\mathbb{C}^{4n\times (\lfloor L/2\rfloor+1)}$ and $W\in\mathbb{R}^{4n\times (\lfloor L/2\rfloor+1)}$ is a learnable real-valued parameter. The softplus transformation constrains the resulting weights to be positive. Consequently, 
\begin{equation}
\angle f(\boldsymbol{X})=\angle\boldsymbol{X}
\end{equation}
Thus, the learned response manipulates spectral magnitudes while preserving the original phase. This restriction keeps the response focused on amplitude modulation and prevents it from introducing arbitrary phase shifts that could alter the temporal alignment of the original signals.

% motivation for the multifilters

However, since the spectral characteristics of proprioceptive histories vary across samples, a single global frequency response may not be sufficient. Therefore, we extend this structure to a mixture of $K$ expert filters.
    
% multifilter definition
Let $\alpha_k\in\mathbb{R}$ denote a gating weight corresponding to the $k$-th expert. The gating weight vector $\boldsymbol{\alpha}=[\alpha_1,\cdots,\alpha_K]\in\mathbb{R}^{K}$ is computed as:
\begin{equation}
    \boldsymbol{\alpha}=\mathrm{softmax}_{\mathrm{e}}\left(\phi(\boldsymbol{x}_{t_h:t}^\delta) \right)
\end{equation}
\begin{equation}
    \phi(\boldsymbol{x}_{t_h:t}^\delta) = \big(\mathrm{flatten}_{4n,L} (\boldsymbol{x}_{t_h:t}^\delta) \big)^T W_p
\end{equation}
$\phi$ is a linear gating layer with a learnable weight $W_p\in\mathbb{R}^{4nL \times K}$ to produce $K$ logits, and the $\mathrm{softmax}_{\mathrm{e}}$ denotes the softmax function applied over the expert dimension. 
The frequency enhancement filter $\mathrm{FEF}(\cdot)$ is then defined as:
\begin{equation}\label{eq:fef}
    \mathrm{FEF}(\boldsymbol{x}_{t_h:t}^\delta)= \mathcal{F}^{-1}\left(\sum_{k=1}^{K} \alpha_k f_k(\boldsymbol{X})\right)
\end{equation}
\noindent
% whats better about the multifilters
where each $f_k$ has its own filter weight $W_k$. This design allows the $\mathrm{FEF}$ to adapt frequency responses to each input sample more flexibly than Eq.~\ref{eq:complexfilt}. 
% dimensionality-invariant
Note that the resulting $\mathrm{FEF}(\boldsymbol{x}_{t_h:t}^\delta)$ has the same dimensionality as $\boldsymbol{x}_{t_h:t}^{\delta}$, thus $\mathrm{FEF}(\cdot)$ is shape-preserving. We then patch-embed $\mathrm{FEF}(\boldsymbol{x}_{t_h:t}^\delta)$ to feed each modality encoder. In our work, we set $K=32$.

\subsection{Loss function}
Reflecting the different estimation forms defined in Section~\ref{subsec:probheads}, we train the two heads with different objectives. We train the trend head by minimizing the mean-squared error of $\hat{\boldsymbol{W}}_{t+1:t_f}^{\mathrm{trend}}$:
\begin{equation}
    \mathcal{L}_{\mathrm{trend}}=\frac{1}{6T}\sum_{i=1}^{6}\sum_{j=1}^{T} \left(\boldsymbol{W}_{i,t+j}^{\mathrm{trend}}-\hat{\boldsymbol{W}}_{i,t+j}^{\mathrm{trend}}\right)^2
\end{equation}
where $i$ and $t+j$ denote channel and time indices. The residual head is trained by minimizing the negative log-likelihood of the ground-truth residual sequence $\boldsymbol{W}^{\mathrm{res}}_{t+1:t_f}$:
\begin{equation}
    \mathcal{L}_{\mathrm{res}}=\frac{1}{6T}\sum_{i=1}^{6}\sum_{j=1}^{T}\frac{1}{2}\left(\frac{(\boldsymbol{W}_{i,t+j}^{\mathrm{res}}-\boldsymbol{\hat{\mu}}^{\mathrm{res}}_{i,t+j})^2}{\exp(\hat{\mathbf{v}}_{i,t+j}^\mathrm{res})}+\hat{\mathbf{v}}_{i,t+j}^{\mathrm{res}}\right)
\end{equation}
The total loss $\mathcal{L}$ is the sum of the trend and residual losses:
\begin{equation}
    \label{eq:loss_total}
    \mathcal{L}=\mathcal{L}_{\mathrm{trend}}+\mathcal{L}_{\mathrm{res}}
\end{equation}
We then minimize $\mathcal{L}$ to optimize the model parameters $\theta$.
We train FDN on a single NVIDIA Tesla V100 GPU with 16~GB of memory, using 12 threads of an Intel Xeon Gold 6136 processor and 64~GB of system memory, running Rocky Linux 9.8 and PyTorch 2.6.0+cu118. Under this configuration, the training takes approximately 14 minutes per run. Inference times are discussed together with the baselines in Section~\ref{subsec:inference_time}.

%% file: src/sections/experiments.tex
\section{Experimental setup}\label{sec:exp}

In this section, we describe the experimental setup for evaluating FDN on real-world data collected from our robotic grinding. The experiments examine how the decomposition-based asymmetric modeling in Section~\ref{sec:fpn} affects accuracy in each frequency band, and whether the multi-step-ahead estimation in Eq.~\ref{eq:model} sacrifices accuracy relative to instantaneous estimation.
% Dataset & cutoffs
We first describe the collected dataset, from which we also determine the cutoff frequencies $f_c$ and $f_c^{\mathrm{dn}}$ that Section~\ref{sec:fpn} assumes to be given.
% Rest
We then describe the data processing, the baselines and training configurations, and the evaluation protocol under time delay.

\subsection{Real-world hydraulic grinding dataset}
\label{subsec:robot_exp}

\begin{figure*}[!tb]
    \centering
    \includegraphics[width=\textwidth]{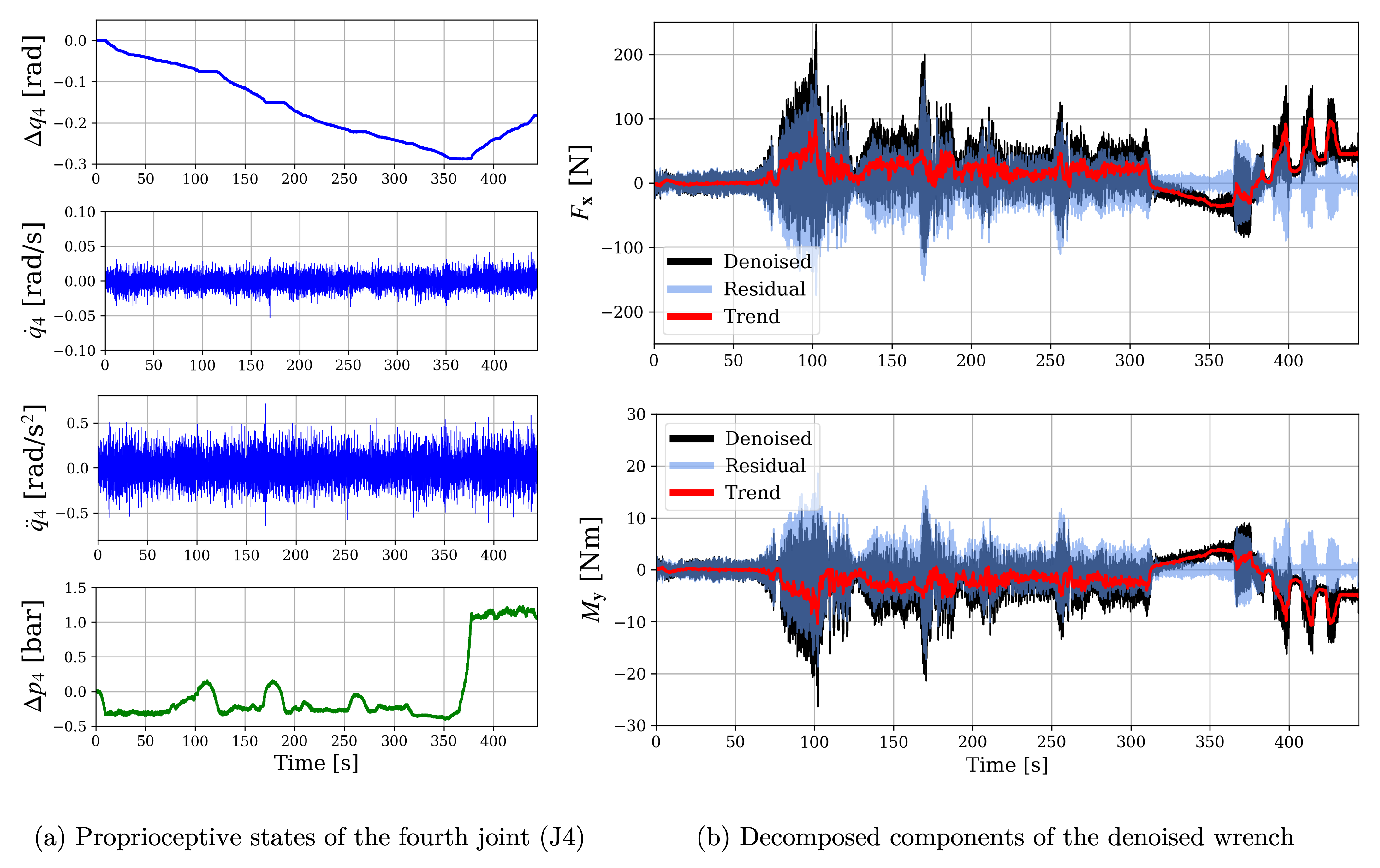}
    \caption{Visualization of the `Stiff-2' episode in the collected hydraulic dataset. (a) Trajectories of $\Delta q_4$, $\dot{q}_4$, $\ddot{q}_4$ and $\Delta p_4$. (b) Denoised and decomposed trajectories of $F_\mathbf{x}$ and $M_\mathbf{y}$ with cutoffs $f_c=1$~Hz and $f_c^{\mathrm{dn}}=15$~Hz.
    }
    \label{fig:dataset_example}
\end{figure*}

\input{src/table/data.tex}

% Overview & control
We conduct real-world robotic grinding teleoperation using a 6-DoF KnR HYDRA-UW3 hydraulic manipulator equipped with a grinder end-effector, as shown in Fig.~\ref{fig:robot_exp}. To mimic a moderate ground excavation task, we grind fixed gypsum blocks along the $\mathbf{-x}$ and $\mathbf{-z}$ directions while constraining translation along the $\mathbf{y}$ axis. 
The objective of each episode is to reach a target horizontal excavation length $d_\mathbf{x}$ and target vertical excavation depth $d_\mathbf{z}$.
During teleoperation, we actuate the intermediate joints $q_2$, $q_3$, and $q_4$, while keeping $q_1$, $q_5$, and $q_6$ fixed. 
The grinder rotates about the $+\mathbf{y}$ axis of the base frame at an idle speed of 10.47~rad/s ($=100$ RPM). The instantaneous horizontal velocity $v_\mathbf{x}$ is bounded within $[-2,0)$~mm/s.

% Sensing & hardware specs
All six joints are actuated by a single hydraulic power unit, with each joint driven by either a revolute or prismatic cylinder mechanism. The joint differential hydraulic pressure $\Delta\boldsymbol{p}$ is measured by the manufacturer's pressure sensing system, which is used as the actuation signal $\boldsymbol{u}$. A 6-axis F/T sensor (HBK MCS10) mounted at the wrist measures the wrench $\boldsymbol{W}$. During manipulation, we collect the six joint angle trajectories $\boldsymbol{q}$ together with $\Delta\boldsymbol{p}$ and $\boldsymbol{W}$ at 100~Hz.

% Data analysis
We collect data over 12 episodes, as summarized in Table~\ref{tbl:rawdata}. $N_{(\cdot)}$ denotes the number of samples in each modality. The episodes were collected in two sessions, denoted as `Soft' and `Stiff'. Each session contains a total of six episodes. We grind softer blocks rapidly in the `Soft' session and stiffer blocks slowly in the `Stiff' session. These session variations resulted in distinct wrench distributions across sessions, as reflected in the maximum force and torque magnitudes. $d_\mathbf{x}$ and $d_\mathbf{z}$ also vary across sessions and episodes. 
In the collected dataset, $d_\mathbf{x}$ ranges from approximately 140 to 236~mm, and $d_\mathbf{z}$ ranges from 10 to 17~mm. The intra-episode average horizontal velocity $\mathbb{E}[v_\mathbf{x}]$ varies from $-1.59$ to $-0.37$~mm/s. The duration of each episode ranges from 149 to 594~s. The total duration is roughly 66~minutes, of which 79\% corresponds to in-contact phases. The collected trajectories and an example wrench decomposition are visualized in Fig.~\ref{fig:dataset_example}.

\subsection{Cutoff frequency selection}
\label{subsec:cutoff_selection}
\begin{figure*}
    \centering
    \includegraphics[width=0.85\textwidth]{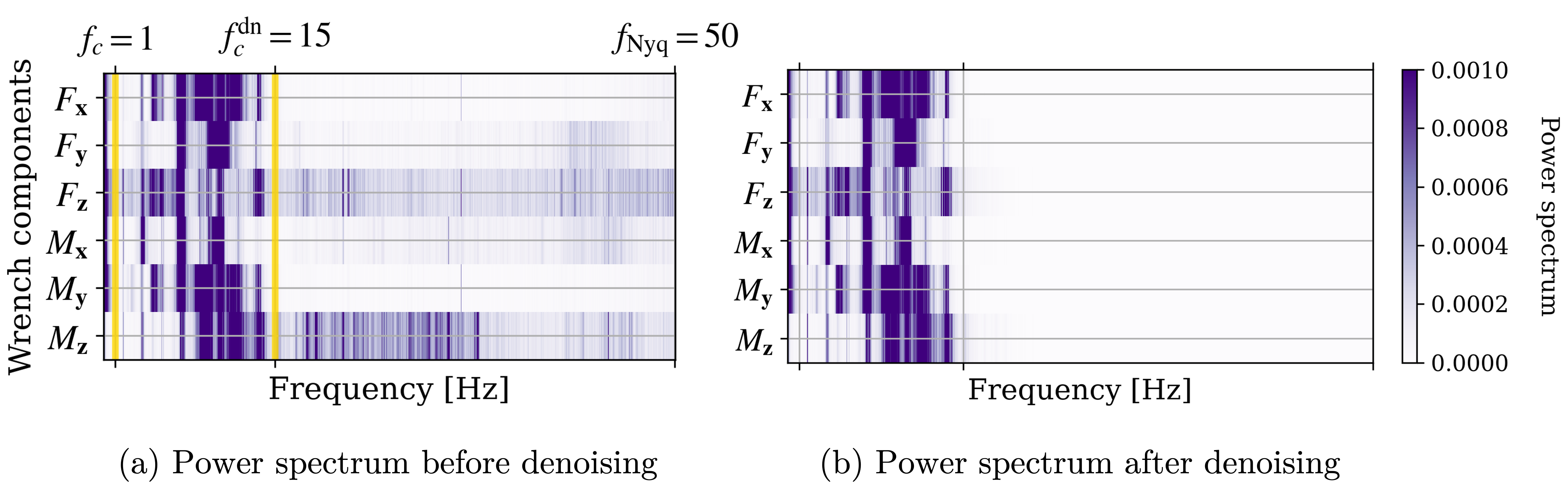}
    \caption{Power spectrum of raw training wrench windows. Each channel is normalized to zero mean and unit variance before the Fourier transform. Power spectrum before denoising (a) and after denoising (b). Yellow vertical lines indicate the decomposition cutoff $f_c=1$~Hz and the denoising cutoff $f_c^{\mathrm{dn}}=15$~Hz applied during training.
    }
    \label{fig:spectral_energy}
\end{figure*}
% Spectral energy of wrench for selecting f_c
We analyze frequency components of the wrench to determine the decomposition and denoising cutoff frequencies $f_c$ and $f_c^{\mathrm{dn}}$. Fig.~\ref{fig:spectral_energy} shows the average power spectrum $|\mathcal{F}(\boldsymbol{W}_{t+1:t+5000})|^2$ of undecomposed, 5,000-step training wrench windows before and after the denoising. We observe that low-frequency energy is concentrated below 1~Hz across all channels, while overall spectral energy is concentrated below 15~Hz. Accordingly, we set $f_c=1$~Hz and $f_c^{\mathrm{dn}}=15$~Hz. This choice of $f_c$ also keeps the trend head focused on the clearly low-frequency-dominant region, while assigning the remaining higher-frequency content to the residual head.

\subsection{Data processing}
\label{subsec:data_processing}

% Data processing
We preprocess the collected data for training. We first transform the wrench from the sensor frame to the base frame and synchronize all modalities to the joint-state timestamps using zero-order hold. We construct $\Delta\boldsymbol{q}$ and $\boldsymbol{q}_0^e$ as in Eq.~\ref{eq:rel_kine}. We align sensing bias in $\Delta\boldsymbol{p}$ and $\boldsymbol{W}$ using the temporal mean of an idle state segment for each channel in each episode. We also set outliers in $\Delta\boldsymbol{p}$ to zero. We then denoise the inputs $\Delta\boldsymbol{q}$ and $\boldsymbol{u}$ using a causal fourth-order Butterworth low-pass filter with cutoff frequency $f_c^{\mathrm{dn}}$. We denoise the output $\boldsymbol{W}$ at the episode level using $\mathrm{FPF}_{\mathrm{low}}$ with $f_c=f_c^{\mathrm{dn}}$ before the horizon-level decomposition in Eq.~\ref{eq:decomp}. We estimate $\boldsymbol{\dot{q}}$ and $\boldsymbol{\ddot{q}}$ from the denoised $\Delta\boldsymbol{q}$ using a causal Savitzky--Golay filter \citep{savitzky1964smoothing} with a window length of 31 and polynomial order of 3, evaluating the derivatives at the endpoint of each trailing window. In particular, fitting a low-order polynomial over this 0.31 s window reduces the noise that numerical differentiation amplifies, while previous works obtain these quantities using cascaded first-order high-pass filters \citep{smith2006neural,dong2020sensorless} or replace the acceleration with its commanded value \citep{pan2023adaptive}.

% Train/Test split
The preprocessed data is split at the episode level to construct training and test datasets. To account for session-dependent wrench distributions, we select two episodes from each session for the test set. From the `Soft' session, we choose the longest (Soft-1) and the shortest (Soft-2) episodes to cover varying trajectory lengths. For the `Stiff' session, we include Stiff-1, which exhibits outlier wrench magnitudes (627 N in $F_\mathbf{x}$ and 193 Nm in $M_\mathbf{z}$, possibly reflecting the deepest penetration $d_{\mathbf{z}}$), and the longest episode (Stiff-3). The test set accounts for 37\% of the total samples. We use the remaining eight episodes ($\approx 63\%$) for training. Training samples are extracted using a temporal stride of 1.

\subsection{Baselines and training configurations}

\label{subsec:num_exp_setup}
% \subsubsection{Setting baselines}
We compare FDN against baselines from robotic force/wrench estimation and time-series forecasting:

    \paragraph{Point-to-point estimators}
    We implement a model-independent neural network (MINN) \citep{smith2006neural} with GELU \citep{hendrycks2016gaussian} activation, as well as a radial basis function (RBF) neural network \citep{chen2019rbf} with 28 basis functions, and channel-independent Gaussian process regression (GPR) \citep{dong2020sensorless} with 1,024 inducing points. These models estimate a current wrench $\boldsymbol{W}_t$ from the current input state vector $\boldsymbol{x}^{\prime}_t=[\boldsymbol{q}_t,\boldsymbol{\dot{q}}_t,\boldsymbol{\ddot{q}}_t,\boldsymbol{u}_t]$.

    \paragraph{Sequence-to-point estimators}
    We implement long short-term memory (LSTM) \citep{kruvzic2022neural} with added layer normalization, and a convolutional neural network (CNN) \citep{pan2024graph} model following the architecture of the original work. These models estimate a current wrench $\boldsymbol{W}_t$ from the input history $\boldsymbol{x}^{\prime}_{t_h:t}$.

    \paragraph{Sequence-to-sequence estimators}
    We implement an autoregressive LSTM encoder-decoder (LSTM-ED) \citep{kao2020exploring} with added layer normalization, a Transformer encoder-decoder with non-autoregressive inference \citep{vaswani2017attention,zhou2021informer}, and modified PatchTST and iTransformer \citep{nie2022time,liu2023itransformer}. Since the original PatchTST and iTransformer models only support endogenous forecasting, where the input and output channels are identical, we incorporate a channel-mixing linear projection and the modified RevIN, which applies inverse normalization to the encoder representation, as described in Section~\ref{subsec:input_repr} to support our input setting. These models estimate a multi-step-ahead wrench $\boldsymbol{W}_{t+1:t_f}$ from the input history $\boldsymbol{x}^{\prime}_{t_h:t}$. In addition, we consider PatchTST-Gaussian, which estimates parameters of time- and channel-independent Gaussian distributions of undecomposed $\boldsymbol{W}_{t+1:t_f}$ from $\boldsymbol{x}^{\prime}_{t_h:t}$ as in Eq.~\ref{eq:residual_heads}. All models except the LSTM-ED estimate $\boldsymbol{W}_{t+1:t_f}$ directly without autoregression.

    We limit the comparison to data-driven estimators. Constructing a model-based estimator for our hydraulic manipulator would require modeling and identifying the cylinder friction of its mixed revolute and prismatic drives and the hydraulic actuation effects, which is the burden that motivates this work in Section~\ref{sec:intro}.

In our experiments, we use $L=100$ and $T=100$, corresponding to estimating one second ahead from one second of history. We set the input joint space dimension $n$ in Eq.~\ref{eq:input_rel} to 6. We normalize the training datasets to zero mean and unit standard deviation. For a controlled comparison, all models are optimized with Adam \citep{kingma2014adam} for 19,330 iterations ($=5$ epochs) using a batch size of 64. Hyperparameters are kept as consistent as possible across models, including latent dimension $D=128$, patch length $P=24$, dropout probability of 0.2, two Transformer encoder layers, and eight attention heads. The remaining model-specific hyperparameters are reported in the released implementation linked in Section~\ref{sec:intro}. All experiments are conducted over three runs, and we report the mean and standard deviation values.

\subsection{Evaluation method}\label{subsec:eval_method}

For evaluation, we assume a constant time delay $t_\mathrm{delay}$ that accounts for estimation and communication during real deployment. Under this setting, each model is used to reconstruct the test episodes by concatenating a single prediction point from each input sample. For point estimators, the estimate produced at time $t$ is compared with the ground truth at time $t+t_\mathrm{delay}$, corresponding to a \textit{delayed zero-order-hold estimate}. For sequence estimators, we extract a prediction point at $t+t_\mathrm{delay}$ from each output sequence and concatenate these points to reconstruct the episodes. The reconstructed episode is then compared directly with the time-aligned ground-truth episode, corresponding to a \textit{multi-step-ahead estimation}.

% Metrics
We then evaluate band-specific accuracies of the reconstructed episodes by decomposing them at the episode level into trends and residuals using $\mathrm{FPF}_{\mathrm{low}}$ with $f_c=1$ Hz. Their episode-level ground truths are also obtained analogously. Then, for the high-frequency band, we measure transient amplitude fidelity by computing the windowed root mean square error ($\mathrm{wRMSE}$) of the residual. Specifically, we compute the root mean square (RMS) values within sliding windows in every test episode, concatenate them, and compute the root mean squared error (RMSE) of the stacked RMS sequence:
\begin{equation}\label{eq:wrmse}
    \mathrm{wRMSE} = \sqrt{ \frac{1}{N_w C}\sum_{t=1}^{N_w}\sum_{c=1}^{C} \big( r_{t,c}(\hat{y}^{\mathrm{res}})-r_{t,c}(y^{\mathrm{res}}) \big)^2 }
\end{equation}
where $C$ is the number of channels, $N_w$ is the number of valid sliding window samples across episodes, and $w$ is the sliding window size. The window-level RMS function $r_{t,c}$ is defined as:
\begin{equation}\label{eq:window_rms}
    r_{t,c}(x) = \sqrt{ \frac{1}{w}\sum_{i=t}^{t+w-1} (x_{i,c})^2 }
\end{equation}
We set $w=10$, corresponding to a 0.1 s window at 100~Hz. For the low-frequency band, we compute the pointwise RMSE of the trends:
\begin{equation}\label{eq:prmse}
    \mathrm{pRMSE} = \sqrt{  \frac{1}{NC}\sum_{i=1}^{N}\sum_{c=1}^{C} (y_{i,c}^{\mathrm{trend}} - \hat{y}_{i,c}^{\mathrm{trend}}  )^2  }
\end{equation}

To additionally evaluate predictions over the full frequency band, we measure the continuous ranked probability score ($\mathrm{CRPS}$), a proper scoring rule widely used for evaluating probabilistic forecasts:

\begin{equation}
    \label{eq:crps}
    \mathrm{CRPS} = \frac{1}{NC}\sum_{i=1}^{N} \sum_{c=1}^{C} \int_{-\infty}^{\infty} \left(  G_{i,c}(x)-\mathbf{1}_{ \{x\geq y_{i,c}\} }   \right)^2 dx
\end{equation}
\noindent
where $G_{i,c}$ is the predicted cumulative distribution function (CDF) and $\mathbf{1}_{ \{x\geq y_{i,c}\}  }$ is the step CDF at observation $y_{i,c}$. The $\mathrm{CRPS}$ reduces to the mean absolute error (MAE) for deterministic models, since $G_{i,c}(x)=\mathbf{1}_{\{ x \geq \hat{y}_{i,c} \}}$.

For FDN and PatchTST-Gaussian, which explicitly parameterize the conditional distribution of the target wrench, their episode-level trend estimates are obtained by decomposing their predictive mean. For FDN, its predictive mean is $\hat{\boldsymbol{W}}^{\mathrm{trend}}+\hat{\boldsymbol{\mu}}^{\mathrm{res}}$. Additionally, we replace the energy $x_{i,c}^2$ in Eq.~\ref{eq:window_rms} with its expectation $\mathbb{E}[x_{i,c}^2]=\mu_{i,c}^2+\sigma_{i,c}^2$. Here, $\mu_{i,c}$ denotes the residual of the predictive mean, and $\sigma_{i,c}$ is the predicted standard deviation. For GPR, we evaluate it as a deterministic point estimator, following \citet{dong2020sensorless}.

%% file: src/table/data.tex
\newcommand{\testrow}{\rowcolor{gray!30}}
\begin{table*}[!tb]
    \begin{center}
        \caption{Overview of the collected hydraulic grinding dataset. For $\max|\boldsymbol{F}|$ and $\max|\boldsymbol{M}|$, the channel with the maximum magnitude is indicated.}
        \label{tbl:rawdata}
        \footnotesize
        \setlength{\tabcolsep}{3pt}
        \begin{tabularx}{\textwidth}{*{2}{>{\centering\arraybackslash}X}*{9}{>{\centering\arraybackslash}X}}
            \hline
            \hline
            Episode & Dataset  & Duration & In-contact & $N_{\boldsymbol{q},\Delta\boldsymbol{p}}$ & $N_{\boldsymbol{W}}$ & $\max|\boldsymbol{F}|$ & $\max|\boldsymbol{M}|$ & $d_\mathbf{x}$  & $d_\mathbf{z}$ & $\mathbb{E} [v_\mathbf{x}]$ \\

            Unit    & -        & [s]      & [s]        & -                                         & -                    & [N]                    & [Nm]                   & [mm]   & [mm]  & [mm/s]             \\

            \hline
            
            Soft-1  & Test     & 289.18   & 214.11     & 28,861                                    & 28,924               & $|F_{\mathbf{x}}|=122$     & $|M_{\mathbf{x}}|=35$      & 158.31 & 11.75 & -0.74              \\
            
            Soft-2  & Test     & 149.34   & 114.15     & 14,909                                    & 14,937               & $|F_{\mathbf{x}}|=96$      & $|M_{\mathbf{y}}|=27$      & 154.70 & 12.58 & -1.36              \\
            Soft-3  & Training & 230.33   & 141.70     & 22,933                                    & 23,035               & $|F_{\mathbf{x}}|=155$     & $|M_{\mathbf{y}}|=47$      & 213.84 & 13.79 & -1.51              \\
            Soft-4  & Training & 152.04   & 108.32     & 15,121                                    & 15,205               & $|F_{\mathbf{z}}|=116$     & $|M_{\mathbf{x}}|=42$      & 160.00 & 12.57 & -1.48              \\
            Soft-5  & Training & 176.63   & 130.39     & 17,618                                    & 17,663               & $|F_{\mathbf{x}}|=134$     & $|M_{\mathbf{y}}|=41$      & 206.74 & 13.23 & -1.59              \\
            Soft-6  & Training & 155.16   & 121.75     & 15,427                                    & 15,517               & $|F_{\mathbf{x}}|=111$     & $|M_{\mathbf{y}}|=31$      & 162.21 & 11.01 & -1.33              \\
            \hline\hline
            
            Stiff-1 & Test     & 418.71   & 336.52     & 41,824                                    & 41,874               & $|F_{\mathbf{x}}|=627$     & $|M_{\mathbf{z}}|=193$     & 235.64 & 16.27 & -0.70              \\
            Stiff-2 & Training & 445.78   & 375.77     & 44,579                                    & 44,581               & $|F_{\mathbf{x}}|=239$     & $|M_{\mathbf{z}}|=69$      & 140.49 & 13.56 & -0.37              \\
            
            Stiff-3 & Test     & 593.85   & 479.60     & 59,366                                    & 59,381               & $|F_{\mathbf{x}}|=364$     & $|M_{\mathbf{z}}|=125$     & 206.57 & 14.61 & -0.43              \\
            Stiff-4 & Training & 396.82   & 324.99     & 39,682                                    & 39,685               & $|F_{\mathbf{x}}|=258$     & $|M_{\mathbf{z}}|=78$      & 157.16 & 13.36 & -0.48              \\
            Stiff-5 & Training & 557.77   & 457.64     & 55,456                                    & 55,756               & $|F_{\mathbf{x}}|=303$     & $|M_{\mathbf{z}}|=120$     & 205.70 & 12.73 & -0.45              \\
            Stiff-6 & Training & 383.12   & 331.67     & 38,306                                    & 38,293               & $|F_{\mathbf{x}}|=260$     & $|M_{\mathbf{z}}|=75$      & 156.05 & 13.03 & -0.47              \\
            \hline
            Total   & -        & 3948.73  & 3136.60    & 394,082                                   & 394,851              & -                      & -                      & -      & -     & -                  \\
            \hline
            \hline
        \end{tabularx}
    \end{center}
\end{table*}

%% file: src/sections/results.tex
\section{Results}\label{sec:results}

We first compare FDN with the baselines under the assumed time delays, and then examine our architectural and input design choices through ablation studies. We also analyze the distribution and dependence structure of the wrench residual $\boldsymbol{W}_{t+1:t_f}^{\mathrm{res}}$ to assess the independent Gaussian modeling in Section~\ref{subsec:probheads}. We finally report the inference speed of each model and explore transferring the learned wrench dynamics from an open-source robot dataset to our setting.

\subsection{Comparison with baselines}
\input{src/table/main_hflf.tex}

Tables~\ref{tbl:hfrmse} and \ref{tbl:lfrmse} report high- and low-frequency band-specific metrics. Notably, FDN and PatchTST-Gaussian exhibit clear improvements in $\mathrm{wRMSE}$, indicating the effectiveness of distribution modeling in the high-frequency band. In particular, FDN demonstrates the best performance in the high-frequency band, reducing $\mathrm{wRMSE}$ by 8\% to 47\% relative to the baselines. For the low-frequency band $\mathrm{pRMSE}$, FDN ranks first or second in the force channels but is comparatively less accurate in the torque channels, remaining competitive overall. In contrast, most baselines exhibit a clear imbalance between the two band-specific metrics. They generally perform well in the low-frequency band, but their accuracy degrades substantially in the high-frequency band. PatchTST-Gaussian shows competitive high-frequency band accuracy by parameterizing the full-band wrench distribution. However, this gain is accompanied by increased error in the low-frequency band, implying that distribution modeling alone is insufficient to improve both bands jointly. FDN effectively mitigates this accuracy trade-off with decomposition-based modeling and enhances the balance between the two bands. Table~\ref{tbl:crps} also shows that FDN achieves the lowest $\mathrm{CRPS}$ across models and time delays, illustrating its superior full-band performance.

\input{src/table/main_crps.tex}

\begin{figure*}[!t]
    \centering
    \includegraphics[width=\textwidth]{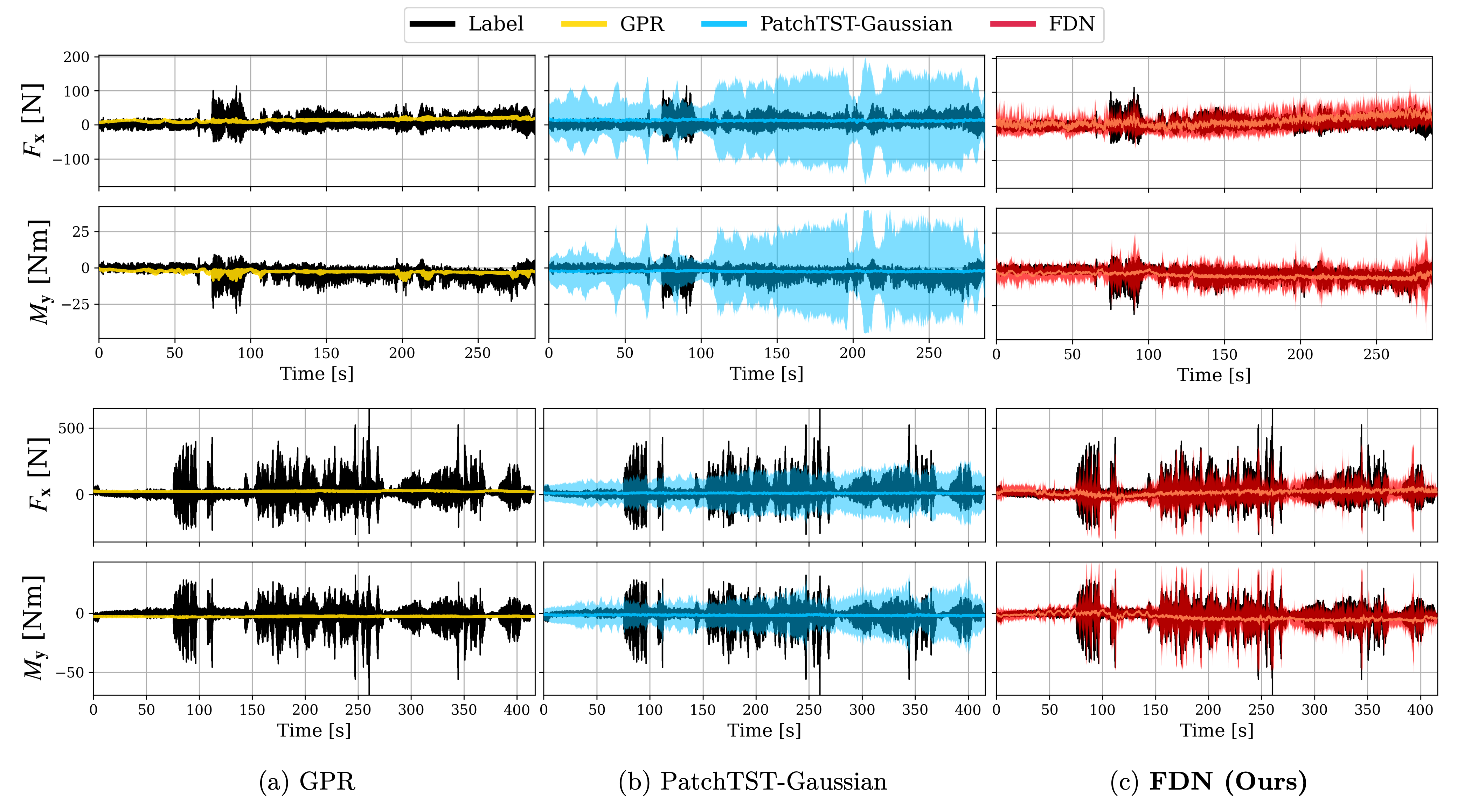}
    \caption{Test episode reconstructions with $t_{\mathrm{delay}}=100$ ms. We visualize the $F_\mathbf{x}$ and $M_\mathbf{y}$, which are representative channels in our experimental setting described in Section~\ref{subsec:robot_exp}. The upper two rows correspond to the `Soft-1' episode, and the lower rows correspond to the `Stiff-1' episode. Colored areas illustrate the prediction interval defined by $\mu\pm3\sigma$.}
    \label{fig:main_result}
\end{figure*}

% Visualize this imbalance and complement FDN
Fig.~\ref{fig:main_result} further visualizes these results using FDN and two representative baselines, GPR and PatchTST-Gaussian. Although GPR is one of our strongest baselines, it shows limited ability to reconstruct high-frequency vibrations, particularly in stiff episodes where wrench magnitudes are larger. PatchTST-Gaussian reconstructs the high-frequency amplitudes through its learned full-band distribution, but its estimated distribution does not accurately capture contact transients. In contrast, FDN successfully captures local high-frequency fluctuations and peaks, together with overall low-frequency trends, which highlights its strong capability for estimating vibration-rich grinding wrench.

\begin{figure*}[!t]
    \centering
    \includegraphics[width=\textwidth]{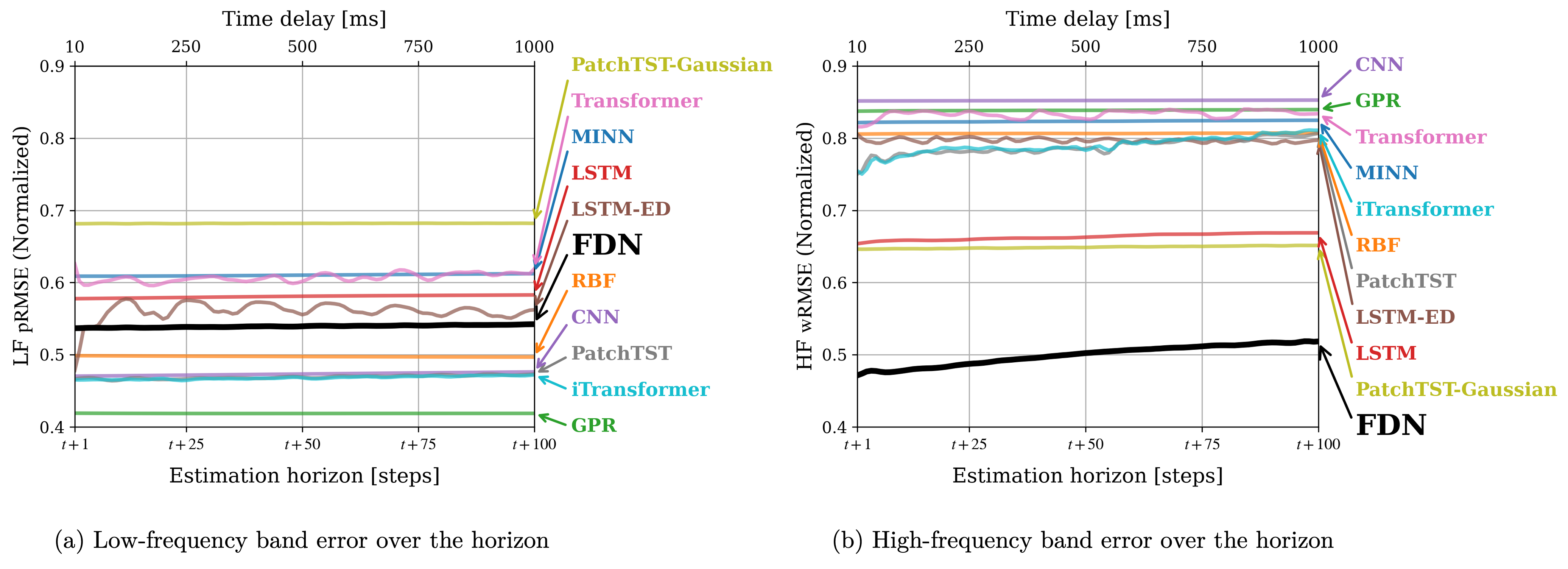}
    \caption{Band-specific errors evaluated at every horizon step from $t+1$ to $t+100$, computed on a normalized scale and aggregated over channels and test episodes. (a) and (b) report $\mathrm{pRMSE}$ and $\mathrm{wRMSE}$, respectively. The upper axis gives the corresponding time delay at the 100~Hz sampling rate, where $t+10$ and $t+100$ are the delays reported in Tables~\ref{tbl:hfrmse} and~\ref{tbl:lfrmse}.}
    \label{fig:horizon_errors}
\end{figure*}

A natural concern with multi-step-ahead estimation is whether temporal extrapolation from the observed history sacrifices accuracy relative to point estimation. Our results show that, under the assumed time delays, multi-step estimators achieve comparable accuracy to, and occasionally outperform, the point estimators across metrics. 
Fig.~\ref{fig:horizon_errors} further shows that this is not specific to the two evaluated delays (100~ms and 1,000~ms) in Tables~\ref{tbl:hfrmse} and~\ref{tbl:lfrmse}. The error levels and their relative ordering across models are maintained over the entire horizon from $t+1$ to $t+100$.
These results support the practical feasibility of multi-step-ahead estimation under the short-horizon assumptions made in Section~\ref{subsec:model}.

% The formulation itself does not guarantee accuracy.
The multi-step-ahead formulation determines whether an estimate is temporally valid after the delay, but not its accuracy. For instance, GPR outperforms most sequence baselines in the low-frequency band despite relying on zero-order-hold estimates, as shown at the bottom of Fig.~\ref{fig:horizon_errors}(a). 
% What matters is the representation
Rather, the accuracy depends more on how the model represents the wrench, since the baselines that do not consider wrench stochasticity do not approach the accuracy of FDN in the high-frequency band.
% Strengthening decomp-based prob modeling
% Rate difference (steeper HF)
For FDN, the two bands degrade at different rates over the horizon, as shown by the steeper increase of $\mathrm{wRMSE}$ in Fig.~\ref{fig:horizon_errors}(b) relative to the nearly flat $\mathrm{pRMSE}$ in Fig.~\ref{fig:horizon_errors}(a). 
% LF are more predictable, aligns with FDN
This observation suggests that the low-frequency content is less sensitive to the horizon and thus more predictable than the high-frequency content, which aligns with our hypothesis of the low-frequency wrench as a deterministic trend and the high-frequency counterpart as a stochastic variation, as discussed in Section~\ref{sec:intro}.

\subsection{Architectural ablation studies}

\input{src/table/ablation.tex}

Table~\ref{tbl:ablation} shows architectural ablation results of FDN. We remove the frequency enhancement filter (w/o FEF), its expert gating weights while retaining multiple filters (w/o FEF-W), its MoE design by reducing to a single unweighted filter (w/o FEF-MoE), the frequency-pass filters in Eq.~\ref{eq:freqpassfilt} (w/o FPF), the modality-specific encoders by replacing them with a single shared encoder for $\boldsymbol{x}_{t_h:t}^\delta$ (w/o ModSpec), and each prediction head (w/o TrdHead and w/o ResHead). When ablating a head, we retain the frequency-pass filter on the remaining output, i.e., $\tilde{\boldsymbol{\mu}}^{\mathrm{res}}$ or $\tilde{\boldsymbol{W}}^{\mathrm{trend}}$, using $\mathrm{FPF}_{\mathrm{low}}$ with $f_c=f_c^{\mathrm{dn}}=15$ Hz. All values are reported on a normalized scale aggregated across channels to remove magnitude effects.

The results indicate that the residual head is the main contributor to modeling the high-frequency band. Removing it, which reduces FDN to a full-band pointwise regression model, increases $\mathrm{wRMSE}$ by 52\%. Removing the trend head, which corresponds to parameterizing a full-band distribution as in PatchTST-Gaussian, also degrades both $\mathrm{wRMSE}$ and $\mathrm{pRMSE}$ by around 20\%, despite a slight reduction in $\mathrm{CRPS}$. Consistent with the previous section, these results show that relying solely on pointwise regression or distribution modeling over the full band is ineffective. Rather, applying asymmetric modeling across frequency bands facilitates band-balanced performance. Replacing the modality-specific encoders with a single encoder also degrades all metrics by around 10\%, showing that the distinct input modalities are better handled with separate encoders.

For the frequency-aware layers, $\mathrm{FEF}$ provides consistent but moderate benefit, as removing it increases $\mathrm{pRMSE}$ by 6.8\% and $\mathrm{wRMSE}$ by 3.4\%. Removing either the expert weighting or the entire MoE structure from the $\mathrm{FEF}$ also degrades the metrics to the level of removing the layer entirely, indicating that both designs are required to retain its benefit. Meanwhile, the quantitative effect of $\mathrm{FPF}$ is marginal, causing less than 1\% change across all metrics. We view it as a design for explicitly imposing our frequency-band prior derived from the ground-truth decomposition in Eq.~\ref{eq:decomp}, rather than a source of improvement.

\subsection{Input ablation studies}
\label{subsec:input_ablation}

\input{src/table/main_input.tex}

We assess the significance of the input formulation in FDN by ablating the input modalities and their corresponding modality-specific encoders. Table~\ref{tbl:input_ablation} shows that all ablated input modalities meaningfully contribute to the final estimation performance, as removing any of them results in a consistent performance drop. Among the individual input modalities, the joint differential pressure $\Delta\boldsymbol{p}$ plays the most critical role, as its ablation causes the largest degradation across all metrics. Its significance also supports our modeling in Section~\ref{subsec:model} that FDN exploits $\Delta\boldsymbol{p}$ as a measurable torque proxy without explicit identification of the pressure--torque mapping.

In Section~\ref{subsec:data_processing}, the joint velocities and accelerations are numerically estimated from the measured position $\boldsymbol{q}$. Such numerical differentiation is sensitive to noise and vibrations and thus can produce inaccurate estimates, especially in vibration-rich settings like ours. Nevertheless, including these estimated quantities consistently improves the performance, particularly in the low-frequency band. Interestingly, $\boldsymbol{\dot{q}}$ and $\boldsymbol{\ddot{q}}$ appear partially redundant in the high-frequency band, as ablating either one has a negligible effect on $\mathrm{wRMSE}$, whereas ablating both increases it by 6.4\%. As expected, retaining only the position variables $\Delta\boldsymbol{q}$ and $\boldsymbol{q}_0^e$ results in the largest performance degradation. These numerical results support our original input definition in Sections~\ref{sec:prelim}~and~\ref{subsec:input_repr}, together with the cross-variable dependence from the dynamics in Eq.~\ref{eq:inv_dynamics}.

\subsection{Residual distribution and dependence analysis}
\label{subsec:correlated_gaussian}

\begin{figure*}[!tb]
    \centering
    \includegraphics[width=0.8\textwidth]{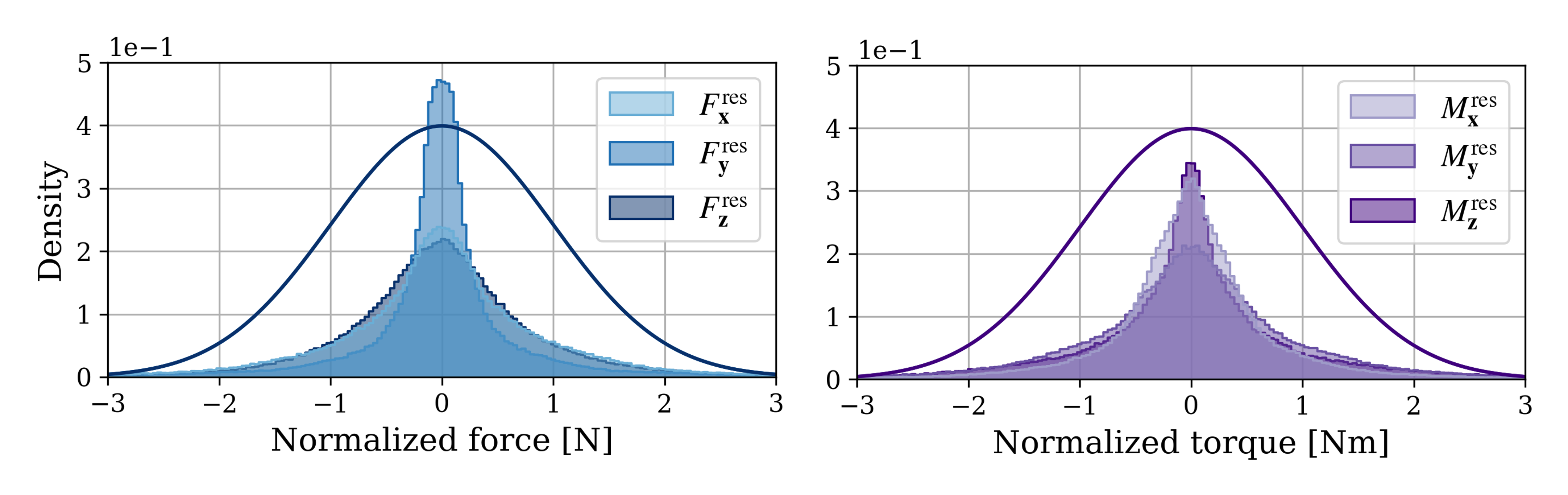}
    \caption{Channel-wise marginal distributions of the normalized training residuals. Each panel overlays histograms of the three force residual channels (Left) and three torque residual channels (Right). Solid lines indicate fitted Gaussian probability density functions.}
    \label{fig:residual_dist}
\end{figure*}

\begin{figure*}[!t]
    \centering
    \includegraphics[width=0.9\textwidth]{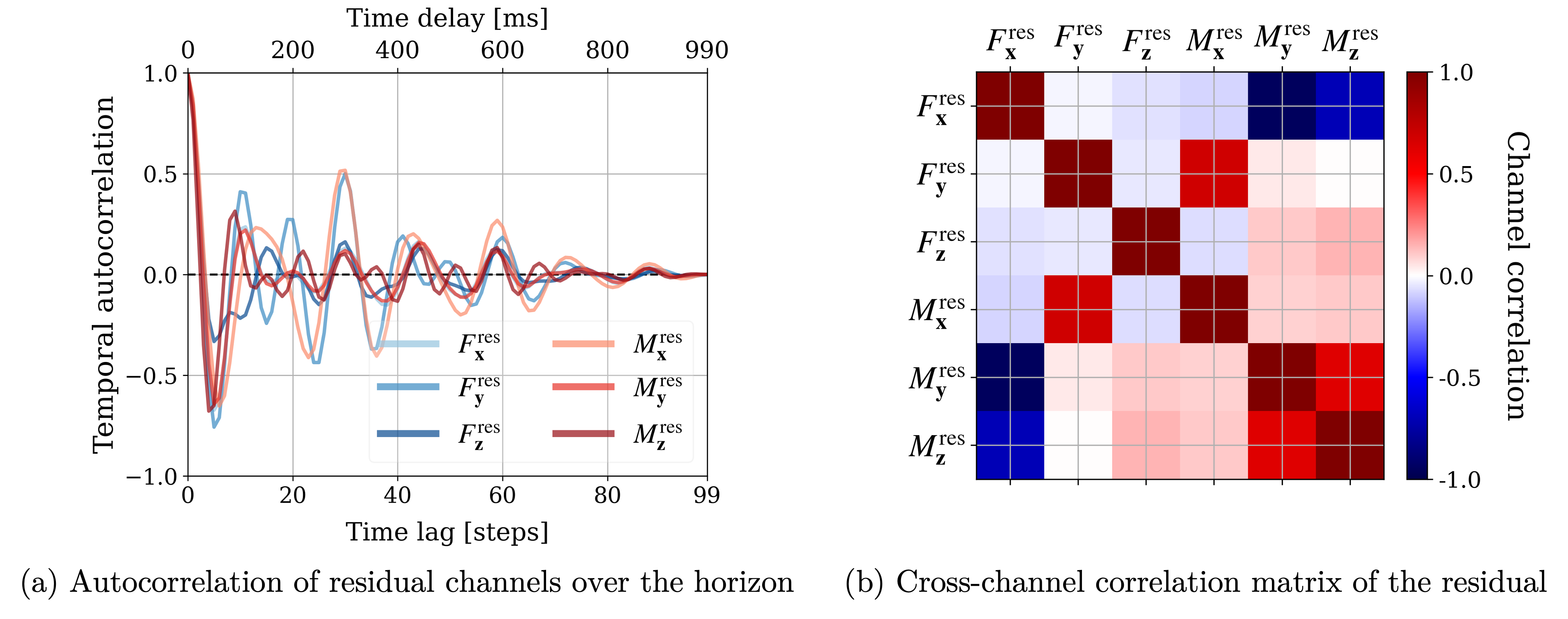}
    \caption{Correlation analysis of the normalized training residuals. (a) Channel-wise temporal autocorrelation functions of the six wrench residual channels over the $T=100$ prediction horizon. (b) Pearson cross-channel correlation matrix averaged over the horizons.}
    \label{fig:residual_corr}
\end{figure*}

The wrench occurring during contact is expected to be correlated across time and channels, which our time- and channel-independent Gaussian modeling in Section~\ref{subsec:probheads} does not represent. However, we find that additionally modeling such correlation for the residual does not improve the performance of FDN.

Figs.~\ref{fig:residual_dist} and~\ref{fig:residual_corr} characterize the marginal distributions and dependency structure of the training residuals. The channel-wise residual marginals are unimodal and approximately symmetric about zero, although their tails are heavier than the fitted Gaussian densities. Nevertheless, the autocorrelation functions and the Pearson correlation matrix reveal oscillatory temporal dependence over the estimation horizon and non-negligible cross-channel dependence, respectively. Consequently, the joint residual distribution does not satisfy the time- and channel-independence adopted in Section~\ref{subsec:probheads}, and these dependencies are not represented by FDN's residual head.

To examine the effect of modeling these dependencies in the residual, we define $D_\sigma=\mathrm{diag}(\mathrm{vec}(\hat{\boldsymbol{\sigma}}^{\mathrm{res}}))\in\mathbb{R}^{TC\times TC}$ and model the residual covariance as:

\begin{equation}
    \label{eq:cov}
    \hat{\Sigma}^{\mathrm{res}}=D_\sigma R D_\sigma
\end{equation}
\noindent
where $R\in\mathbb{R}^{TC\times TC}$ is the correlation matrix. Assuming a Kronecker-structured correlation \citep{werner2008estimation}, we parameterize $R$ as:
\begin{equation}
    R=R_T\otimes R_C
\end{equation}
where $R_T\in\mathbb{R}^{T\times T}$ and $R_C\in\mathbb{R}^{C\times C}$ denote temporal and cross-channel correlation matrices, respectively, and $\otimes$ is the Kronecker product. We implement three FDN variations which additionally learn (1) channel correlation $R_C$ shared across time steps with a fixed $R_T=I_T$, (2) temporal correlation $R_T$ shared across channels with a fixed $R_C=I_C$, and (3) time-channel coupled correlation $R_T\otimes R_C$. These variants learn the correlation matrix components through Cholesky factors \citep{hildebrand1987introduction} initialized to identity. For the time- and channel-independent Gaussian residual in the original FDN, $R=I$ is fixed.

\input{src/table/mvn.tex}

Despite the enhanced statistical expressiveness, Table~\ref{tbl:mvn} shows that none of these variants improve the band-specific estimation performance. Particularly in HF-$\mathrm{wRMSE}$, the performance degradation tends to be more evident in models with larger degrees of freedom in $R$. This trend implies that learning the extended covariance structure may increase optimization complexity without providing additional performance benefit. Here, we omit the $\mathrm{CRPS}$ used in our previous evaluations since its marginal form in Eq.~\ref{eq:crps} does not account for the correlated structures. Consequently, we retain the time- and channel-independent Gaussian in FDN's residual head as a practical implementation.

\subsection{Inference speed}\label{subsec:inference_time}

\input{src/table/main_walltime.tex}

Table~\ref{tbl:walltime} reports the CPU inference wall time of each model. The reported times are averaged over 1,000 forward passes with a batch size of one. We measure inference time using a single CPU thread on an Intel Xeon Gold 6130 processor with 64~GB of system memory, running Rocky Linux 9.8 and PyTorch 2.6.0+cu118. We observe that FDN has a moderate inference speed, with higher latency than the non-autoregressive sequence baselines—Transformer \citep{vaswani2017attention,zhou2021informer}, PatchTST \citep{nie2022time} and its Gaussian variant, and iTransformer \citep{liu2023itransformer}—but lower latency than the autoregressive baseline, LSTM-ED \citep{kao2020exploring}. Note that the latency does not scale with the model size, as FDN has around four times more parameters than PatchTST while requiring about twice its inference time. Point estimators are generally faster, while GPR requires substantially longer inference time because of the large number of inducing points (1,024) used in our implementation. This cost could be reduced by using fewer inducing points, although this would introduce an accuracy--compute trade-off. Importantly, all models complete a forward pass in less than 17~ms on average, which is substantially shorter than the delay assumed in our evaluation (100~ms and 1,000~ms). Further latency reductions may be possible through parallelization, GPU utilization, and hardware- or architecture-specific optimization.

\subsection{Transfer learning studies}
\label{subsec:transfer}
% Motivation and how to pretrain
 As an exploratory study, we pretrain the model on RH20T \citep{fang2024rh20t}, a large-scale open-source robot dataset containing proprioception and wrench data obtained from teleoperated everyday manipulation. We attempt to learn \textit{proprioception-to-wrench dynamics} from RH20T and transfer them to our downstream hydraulic grinding scenario. Importantly, RH20T provides joint torque $\boldsymbol{\tau}$ measured from motor-actuated robots, whereas our downstream setting measures joint differential pressure $\Delta \boldsymbol{p}$ of the hydraulic joints. This modality gap in actuation signals $\boldsymbol{u}$ should be considered in the transfer learning pipeline to avoid negative transfer caused by actuation mismatch.

 \subsubsection{Pretraining data processing}
 We first extract 11,131 valid episodes from RH20T. Then, within each selected episode, we linearly interpolate $\boldsymbol{q}$ sampled at 10~Hz to the 100~Hz timestamps of $\boldsymbol{W}$. We subsequently follow the same data processing pipeline in Section~\ref{subsec:data_processing}, including relative position construction, outlier removal, episode-level denoising, horizon-level decomposition, and derivative estimation. We skip only the sensing bias alignment step here. We use the same decomposition and denoising cutoffs $f_c=1$~Hz and $f_c^{\mathrm{dn}}=15$~Hz for consistency. We extract training samples using a temporal stride of 10 due to the large number of samples in RH20T.

\subsubsection{Proprioception-to-wrench transfer}
\label{subsubsec:pretrain}
We pretrain on RH20T to learn proprioception-to-wrench dynamics of everyday object manipulation. To this end, we use the processed $\boldsymbol{q}$ and $\boldsymbol{W}$ trajectories and numerical derivatives of $\boldsymbol{q}$, while retaining the same input, output and loss definitions in Section~\ref{sec:fpn}. Since the dataset covers both 6- and 7-DoF robots, we set $n=7$ to construct a 35-dimensional input vector and set the 7th-joint channels $\Delta q_7, \dot{q}_7, \ddot{q}_7$, and $q_{0,7}^e$ to zero in $\boldsymbol{x}_{t_h:t}$ for 6-DoF samples. 

To address the modality gap between $\boldsymbol{\tau}$ and $\Delta\boldsymbol{p}$ across the datasets, we set actuation channels in $\boldsymbol{x}_{t_h:t}^\delta$ to zero before $\mathrm{FEF}$, and the subsequent encoder representation $z_{\boldsymbol{u}}$ also to zero to ignore the effect of $\boldsymbol{u}$ during pretraining. Correspondingly, the parameters of $\mathrm{Enc}_{\boldsymbol{u}}$ and its embedding layer are kept frozen. We also skip RevIN for the zeroed channels.

After the pretraining, we transfer the learned proprioception-to-wrench dynamics to the downstream by initializing $\mathrm{Enc}_{\boldsymbol{\Delta q}}$, $\mathrm{Enc}_{\boldsymbol{\dot{q}}}$, $\mathrm{Enc}_{\boldsymbol{\ddot{q}}}$ and their embedding layers, and $\mathrm{Enc}_{\boldsymbol{q}_0^e}$ in Fig.~\ref{fig:model} with the pretrained parameters, while initializing the other layers from scratch. We normalize the downstream inputs with the RH20T training statistics. In the downstream, we first perform linear probing for 5 epochs while freezing the pretrained parameters, and then unfreeze all parameters to fine-tune the model end-to-end for another 5 epochs.

\subsubsection{Transfer results and analysis}

\input{src/table/pretrain.tex}

Table~\ref{tbl:pretrain} summarizes transfer learning results of our pretraining under a fixed budget of 100K training iterations. Here, we vary the data utilization ratios from 20\% to 100\% of the 11,131 pretraining episodes. We also assess the effect of input formulations on transfer by comparing absolute position inputs (`A') with the relative position inputs (`R') in Eq.~\ref{eq:rel_kine}. When adopting the absolute position formulation, we remove the initial position encoder $\mathrm{Enc}_{\boldsymbol{q_0^e}}$ from FDN.

Under the relative position formulation, fine-tuning generally outperforms linear probing, indicating that end-to-end adaptation is required to realize transfer gains. In particular, at 60\% utilization, fine-tuning with relative positions improves $\mathrm{pRMSE}$ and $\mathrm{CRPS}$ by around 8\% and 4\% over training from scratch, while slightly degrading $\mathrm{wRMSE}$ by less than 1\%. In contrast, the absolute position formulation neither provides consistent transfer gains nor outperforms the relative position formulation, indicating that aligning episode-wise position offsets improves transferability in our setting.

\begin{figure*}[!tb]
    \centering
    \includegraphics[width=0.65\textwidth]{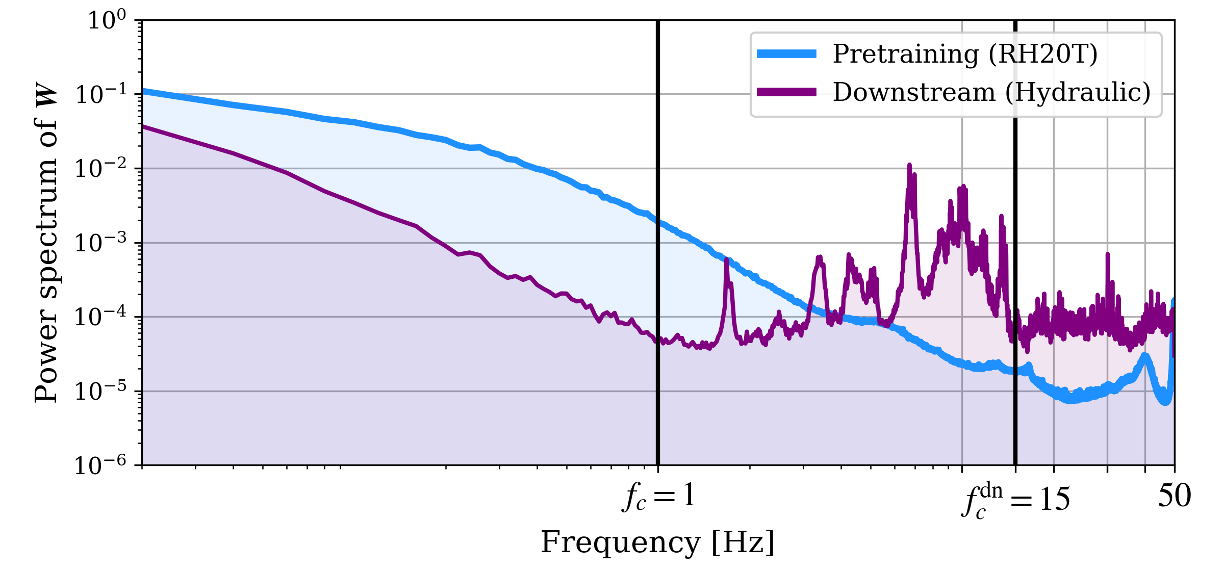}
    \caption{Channel-averaged power spectrum of $\boldsymbol{W}$ in the pretraining (RH20T) and downstream (Hydraulic) datasets, computed from the training splits only. Each channel is normalized to zero mean and unit variance before the Fourier transform, and the spectra are shown without denoising to display the full frequency content. Both axes are logarithmic. Vertical lines mark the cutoffs $f_c=1$~Hz and $f_c^{\mathrm{dn}}=15$~Hz applied during training. }
    \label{fig:energy_comparison}
\end{figure*}

We further interpret our results with the physical properties of the datasets summarized in Table~\ref{tbl:datasets_comparison} and the power spectrum of wrench measurements in Fig.~\ref{fig:energy_comparison}. RH20T is strongly low-frequency dominant, whereas the downstream hydraulic wrench is dominated by high-frequency content. This spectral mismatch reflects the physical differences between the two settings, mainly in the task and actuation. Nevertheless, the two datasets are expected to retain shared dynamics since both measure contact wrenches of teleoperated arm manipulators at their wrists. Our transfer results imply that some of the shared components in the low-frequency dynamics are transferable across the two heterogeneous robot datasets. Conversely, the high-frequency dynamics of the downstream wrench appear to be more domain-specific, consistent with the spectral mismatch between the datasets and the limited transfer gain in $\mathrm{wRMSE}$. 

% utilization analysis
Across utilization ratios, we find the best transfer performance at 60\% utilization under the relative position formulation. Since the effective number of epochs decreases as the amount of pretraining data increases under the fixed iterations, this result should be interpreted as a budget-dependent optimum.

%% file: src/table/main_hflf.tex
\begin{table*}[tb!]
    \caption{High-frequency windowed RMS errors of models.}
    \label{tbl:hfrmse}
    \footnotesize
    \begin{center}
        \begin{tabularx}{\textwidth}{
    >{\centering\arraybackslash}X>{\centering\arraybackslash}X
    *{2}{>{\centering\arraybackslash}X}
    *{2}{>{\centering\arraybackslash}X}
}
            \hline
            \hline
             \multirow{3}{*}{Multi-step-ahead}&  & \multicolumn{4}{c}{$\mathrm{wRMSE}$, High-frequency Windowed RMS Error}                                                                         \\
            \cline{2-6}
            & Time Delay & \multicolumn{2}{c}{100 ms} & \multicolumn{2}{c}{1,000 ms}                                          \\
            & Force/Torque Unit & [N] & [Nm] & [N] & [Nm]              \\
            \hline

            \multirow{5}{*}{\faTimes} & MINN & 24.359 $\pm$ 0.035 & 4.909 $\pm$ 0.003 & 24.431 $\pm$ 0.032 & 4.925 $\pm$ 0.003 \\

             & RBF & 24.130 $\pm$ 0.110 & 4.753 $\pm$ 0.018 & 24.164 $\pm$ 0.109 & 4.759 $\pm$ 0.017  \\

             & GPR & 24.974 $\pm$ 0.004 & 4.958 $\pm$ 0.001 & 25.013 $\pm$ 0.004 & 4.970 $\pm$ 0.001 \\

             & LSTM & 18.592 $\pm$ 0.237 & 3.867 $\pm$ 0.046 & 19.110 $\pm$ 0.211 & 3.938 $\pm$ 0.043 \\ 

             & CNN & 25.105 $\pm$ 0.017 & 5.098 $\pm$ 0.003 & 25.140 $\pm$ 0.017 & 5.104 $\pm$ 0.003 \\ 
            \hline
            \multirow{6}{*}{\faCheck} & LSTM-ED & 23.262 $\pm$ 0.456 & 4.683 $\pm$ 0.110 & 23.213 $\pm$ 0.474 & 4.659 $\pm$ 0.134 \\ 

             & Transformer & 24.732 $\pm$ 0.073 & 4.996 $\pm$ 0.019  & 24.669 $\pm$ 0.073 & 4.979 $\pm$ 0.024 \\ 

             & PatchTST & 22.990 $\pm$ 0.057 & 4.673 $\pm$ 0.011 & 23.777 $\pm$ 0.044 & 4.829 $\pm$ 0.008  \\ 

             & PatchTST-Gaussian & \underline{15.710} $\pm$ 0.080 & \underline{3.435} $\pm$ 0.030 & \underline{16.022} $\pm$ 0.039 & \underline{3.469} $\pm$ 0.026 \\ 

             & iTransformer & 22.708 $\pm$ 0.103 & 4.624 $\pm$ 0.007 & 23.851 $\pm$ 0.107 & 4.850 $\pm$ 0.014 \\ 

            \cline{2-6}
             & FDN & \textbf{13.284} $\pm$ 0.176 & \textbf{2.664} $\pm$ 0.021 & \textbf{14.721} $\pm$ 0.094 & \textbf{2.969} $\pm$ 0.032 \\ 

            \hline\hline
        \end{tabularx}
    \end{center}
\end{table*}

\begin{table*}[tb!]
    \caption{Low-frequency pointwise RMSE of models.}
    \label{tbl:lfrmse}
    \footnotesize
    \begin{center}
        % \begin{tabular}{c|ccccccccc}
        \begin{tabularx}{\textwidth}{
    >{\centering\arraybackslash}X>{\centering\arraybackslash}X
    *{2}{>{\centering\arraybackslash}X}
    *{2}{>{\centering\arraybackslash}X}
}
            \hline
            \hline
            \multirow{3}{*}{Multi-step-ahead} &  & \multicolumn{4}{c}{$\mathrm{pRMSE}$, Low-frequency Pointwise RMSE}                                                                        \\
            \cline{2-6}
            & Time Delay & \multicolumn{2}{c}{100 ms} & \multicolumn{2}{c}{1,000 ms}                                         \\
            & Force/Torque Unit & [N] & [Nm] & [N] & [Nm]              \\
            \hline

            \multirow{5}{*}{\faTimes} & MINN & 12.230 $\pm$ 0.494 & 3.612 $\pm$ 0.095  & 12.274 $\pm$ 0.489 & 3.623 $\pm$ 0.093 \\

             & RBF & 10.455 $\pm$ 0.256 & 3.076 $\pm$ 0.109 & 10.432 $\pm$ 0.265 & 3.067 $\pm$ 0.109 \\

             & GPR & \underline{9.362} $\pm$ 0.016 & \textbf{2.285} $\pm$ 0.019 & \textbf{9.362} $\pm$ 0.018 & \textbf{2.281} $\pm$ 0.019  \\

             & LSTM & 13.040 $\pm$ 1.102 & 3.289 $\pm$ 0.192  & 13.189 $\pm$ 1.029 & 3.300 $\pm$ 0.189 \\

             & CNN & 10.322 $\pm$ 0.496 & \underline{2.485} $\pm$ 0.071  & 10.765 $\pm$ 0.457 & \underline{2.511} $\pm$ 0.072 \\ 
            \hline

            \multirow{6}{*}{\faCheck} & LSTM-ED & 11.873 $\pm$ 1.005 & 3.199 $\pm$ 0.180  & 11.889 $\pm$ 1.433 & 3.231 $\pm$ 0.278  \\

             & Transformer & 10.939 $\pm$ 1.326 & 2.954 $\pm$ 0.182  & 11.009 $\pm$ 1.101 & 3.023 $\pm$ 0.168 \\ 

             & PatchTST & 9.898 $\pm$ 0.018 & 2.545 $\pm$ 0.007  & 10.223 $\pm$ 0.038 & 2.578 $\pm$ 0.006 \\ 

             & PatchTST-Gaussian & 13.562 $\pm$ 0.028 & 4.358 $\pm$ 0.016 & 13.563 $\pm$ 0.023 & 4.352 $\pm$ 0.015 \\ 

             & iTransformer & 9.994 $\pm$ 0.094 & 2.587 $\pm$ 0.023 & 10.326 $\pm$ 0.102 & 2.612 $\pm$ 0.022 \\ 
            
             \cline{2-6}
             & FDN & \textbf{9.102} $\pm$ 0.202 & 3.825 $\pm$ 0.045 & \underline{9.411} $\pm$ 0.180 & 3.830 $\pm$ 0.044 \\ 

            \hline\hline
        \end{tabularx}
    \end{center}
\end{table*}

%% file: src/table/main_crps.tex
\begin{table*}[!tb]
    \caption{Continuous ranked probability scores of models.}
    \label{tbl:crps}
    \footnotesize
    \begin{center}
        \begin{tabularx}{\textwidth}{
    >{\centering\arraybackslash}X>{\centering\arraybackslash}X
    *{2}{>{\centering\arraybackslash}X}
    *{2}{>{\centering\arraybackslash}X}
}
            \hline
            \hline
            \multirow{3}{*}{Multi-step-ahead} &  & \multicolumn{4}{c}{$\mathrm{CRPS}$}                                                                        \\
            \cline{2-6}
            & Time Delay & \multicolumn{2}{c}{100 ms} & \multicolumn{2}{c}{1,000 ms}                                         \\
            & Force/Torque Unit & [N] & [Nm] & [N] & [Nm]              \\
            \hline

            \multirow{5}{*}{\faTimes} & MINN & 13.006 $\pm$ 0.100 & 3.900 $\pm$ 0.038 & 13.029 $\pm$ 0.101 & 3.905 $\pm$ 0.038 \\

             & RBF & 12.561 $\pm$ 0.134 & 3.855 $\pm$ 0.063 & 12.574 $\pm$ 0.142 & 3.856 $\pm$ 0.064 \\ 

             & GPR & 12.095 $\pm$ 0.005 & 3.520 $\pm$ 0.007  & 12.124 $\pm$ 0.005 & 3.528 $\pm$ 0.007 \\ 

             & LSTM & 14.622 $\pm$ 0.518 & 4.332 $\pm$ 0.029 & 14.847 $\pm$ 0.581 & 4.368 $\pm$ 0.058 \\ 

             & CNN & 12.772 $\pm$ 0.144 & 3.727 $\pm$ 0.046 & 12.848 $\pm$ 0.142 & 3.739 $\pm$ 0.047 \\ 
            \hline

            \multirow{6}{*}{\faCheck} & LSTM-ED & 13.384 $\pm$ 0.472 & 4.064 $\pm$ 0.147 & 13.436 $\pm$ 0.525 & 4.064 $\pm$ 0.161 \\ 

             & Transformer & 12.888 $\pm$ 0.299 & 3.902 $\pm$ 0.102 & 12.955 $\pm$ 0.228 & 3.941 $\pm$ 0.084 \\ 

             & PatchTST & 12.259 $\pm$ 0.025 & 3.810 $\pm$ 0.000 & 12.425 $\pm$ 0.020 & 3.827 $\pm$ 0.005 \\ 

             & PatchTST-Gaussian & \underline{9.782} $\pm$ 0.014 & \underline{3.279} $\pm$ 0.009 & \underline{9.802} $\pm$ 0.011 & \underline{3.276} $\pm$ 0.008 \\ 

             & iTransformer & 12.213 $\pm$ 0.061 & 3.836 $\pm$ 0.009 & 12.342 $\pm$ 0.034 & 3.830 $\pm$ 0.008 \\ 

            \cline{2-6}
             & FDN & \textbf{9.083} $\pm$ 0.088 & \textbf{3.256} $\pm$ 0.029 & \textbf{9.156} $\pm$ 0.086 & \textbf{3.262} $\pm$ 0.032 \\ 

            \hline\hline
        \end{tabularx}
    \end{center}
\end{table*}

%% file: src/table/ablation.tex
\begin{table*}[!tb]
    \caption{Architectural ablation results. Metrics are aggregated on the normalized scale.
    }
    \label{tbl:ablation}
    \footnotesize
    \begin{center}
        \begin{tabularx}{\textwidth}{c*{2}{>{\centering\arraybackslash}X}>{\centering\arraybackslash}X}
            \hline\hline
             & HF $\mathrm{wRMSE}$ & LF $\mathrm{pRMSE}$ & $\mathrm{CRPS}$             \\
            \hline

            w/o FEF & 0.518 $\pm$ 0.002 (+3.4\%) & 0.569 $\pm$ 0.006 (+6.8\%) & 0.534 $\pm$ 0.003 (+3.1\%) \\ 

            w/o FEF-W &  0.518 $\pm$ 0.003 (+3.4\%) & 0.572 $\pm$ 0.009 (+7.3\%) & 0.534 $\pm$ 0.005 (+3.1\%) \\ 

            w/o FEF-MoE & 0.515 $\pm$ 0.002 (+2.8\%) & 0.565 $\pm$ 0.000 (+6.0\%) & 0.531 $\pm$ 0.001 (+2.5\%) \\ 

            w/o FPF & \textbf{0.500} $\pm$ 0.001 (-0.2\%) & 0.537 $\pm$ 0.012 (+0.8\%) & 0.520 $\pm$ 0.005 (+0.4\%) \\ 

            w/o ModSpec & 0.553 $\pm$ 0.002 (+10.4\%) & 0.606 $\pm$ 0.001 (\underline{+13.7\%}) & 0.559 $\pm$ 0.001 (\underline{+7.9\%}) \\ 

            w/o TrdHead & 0.615 $\pm$ 0.049 (\underline{+22.8\%}) & 0.631 $\pm$ 0.063 (\textbf{+18.4\%}) & \textbf{0.513} $\pm$ 0.013 (-1.0\%) \\ 

            w/o ResHead & 0.765 $\pm$ 0.001 (\textbf{+52.7\%}) & 0.537 $\pm$ 0.007 (+0.8\%) & 0.675 $\pm$ 0.004 (\textbf{+30.3\%}) \\ 

            \hline

            FDN & \underline{0.501} $\pm$ 0.001 & \textbf{0.533} $\pm$ 0.006 & \underline{0.518} $\pm$ 0.004 \\

            \hline\hline
        \end{tabularx}
    \end{center}
\end{table*}

%% file: src/table/main_input.tex
\begin{table*}
    \caption{Effect of input variables on estimation performance. Metrics are aggregated on the normalized scale.}
    \label{tbl:input_ablation}
    \footnotesize
    \begin{center}
        \begin{tabularx}{\textwidth}{>{\centering\arraybackslash}c
    *{3}{>{\centering\arraybackslash}X}
}
            \hline
            \hline                                        
            Ablated Inputs & HF $\mathrm{wRMSE}$ & LF $\mathrm{pRMSE}$ & $\mathrm{CRPS}$ \\

            \hline
            
            $\boldsymbol{\dot{q}}$
            & 0.500 $\pm$ 0.004 (+0.2\%)
            & 0.583 $\pm$ 0.014 (+9.6\%)
            & 0.542 $\pm$ 0.007 (+4.6\%) \\
            
            $\boldsymbol{\ddot{q}}$
            & \textbf{0.499} $\pm$ 0.002 (+0.0\%)
            & 0.574 $\pm$ 0.026 (+7.9\%)
            & 0.538 $\pm$ 0.013 (+3.9\%) \\
            
            $\boldsymbol{\dot{q}},\boldsymbol{\ddot{q}}$
            & 0.531 $\pm$ 0.002 (+6.4\%)
            & 0.587 $\pm$ 0.041 (+10.3\%)
            & 0.539 $\pm$ 0.024 (+4.1\%) \\
            
            $\boldsymbol{u} (=\Delta\boldsymbol{p})$
            & 0.572 $\pm$ 0.003 (+14.6\%)
            & 0.604 $\pm$ 0.003 (+13.5\%)
            & 0.554 $\pm$ 0.002 (+6.9\%) \\

            $\boldsymbol{\dot{q}}, \boldsymbol{\ddot{q}},\boldsymbol{u}$ 
            & 0.701 $\pm$ 0.050 (+40.5\%)
            & 0.666 $\pm$ 0.014 (+25.2\%)
            & 0.583 $\pm$ 0.007 (+12.5\%) \\

            \hline
            $\varnothing$
            & \textbf{0.499} $\pm$ 0.003
            & \textbf{0.532} $\pm$ 0.007
            & \textbf{0.518} $\pm$ 0.002 \\
            
            \hline\hline
        \end{tabularx}
    \end{center}
\end{table*}

%% file: src/table/mvn.tex
\begin{table*}
    \caption{Effect of residual correlation structure on band-specific estimation performance. Metrics are aggregated on the normalized scale.}
    \label{tbl:mvn}
    \footnotesize
    \begin{center}
        \begin{tabularx}{\textwidth}{>{\centering\arraybackslash}c
    *{3}{>{\centering\arraybackslash}X}
}
            \hline
            \hline                                        
             & HF $\mathrm{wRMSE}$ & LF $\mathrm{pRMSE}$ & Correlation DoF\\

            \hline
            $I_T\otimes R_C$ (Channel) 
            & 0.533 $\pm$ 0.004 (+6.8\%) 
            & 0.555 $\pm$ 0.020 (+4.3\%) 
            & 15\\

            $R_T\otimes I_C$ (Time)  
            & 0.602 $\pm$ 0.003 (+20.6\%) 
            & 0.551 $\pm$ 0.018 (+3.6\%) 
            & 4,950\\

            $R_T\otimes R_C$ (Time-Channel) 
            & 0.638 $\pm$ 0.005 (+27.9\%) 
            & 0.554 $\pm$ 0.018 (+4.1\%) 
            & 4,965\\

                \hline
            $I$ (Independent) 
            & \textbf{0.499} $\pm$ 0.003 
            & \textbf{0.532} $\pm$ 0.007 
            & 0 \\
            \hline\hline
        \end{tabularx}
    \end{center}
\end{table*}

%% file: src/table/main_walltime.tex
\begin{table*}
    \caption{CPU inference time and model sizes.}
    \label{tbl:walltime}
    \footnotesize
    \begin{center}
        \begin{tabularx}{\textwidth}{>{\centering\arraybackslash}X
    *{3}{>{\centering\arraybackslash}X}
}
            \hline
            \hline                                        
            & Wall Time [ms] & Relative & Optimized Parameters \\

            \hline
            MINN & 0.038 $\pm$ 0.003 & 0.004 & 3,974 \\
            RBF & 0.062 $\pm$ 0.004 & 0.006 & 868 \\
            GPR & 16.663 $\pm$ 0.737 & 1.582 & 3,302,563 \\
            LSTM & 6.229 $\pm$ 0.423 & 0.591 & 80,902 \\
            CNN & 5.936 $\pm$ 0.226 & 0.564 & 81,766 \\
            LSTM-ED & 13.186 $\pm$ 0.520 & 1.252 & 151,814 \\
            Transformer & 4.373 $\pm$ 0.398 & 0.415 & 680,838 \\
            PatchTST & 5.475 $\pm$ 0.180 & 0.520 & 470,266 \\
            PatchTST-Gaussian & 5.482 $\pm$ 0.236 & 0.520 & 534,366 \\
            iTransformer & 1.114 $\pm$ 0.036 & 0.106 & 422,778 \\
            \hline
            FDN & 10.533 $\pm$ 0.413 & \textbf{1.000} & 1,950,178 \\
            
            \hline\hline
        \end{tabularx}
    \end{center}
\end{table*}

%% file: src/table/pretrain.tex
\begin{sidewaystable*}[p]
    \centering
    \begin{minipage}{0.81\textheight}
    \caption{Transfer learning results. LP refers to linear probing, and FT refers to fine-tuning. A and R denote absolute and relative position inputs, respectively. 0\% corresponds to training from scratch. Metrics are aggregated on the normalized scale. All pretraining iterations are fixed to 100K steps.
    }
\label{tbl:pretrain}
    \footnotesize
    \begin{center}
    \renewcommand{\arraystretch}{1.6}
\begin{tabular}{cc*{6}{c}}
                \hline
                \hline
                \multicolumn{2}{c}{Pretraining Data Util.}&0\%&20\%&40\%&60\%&80\%&100\%\\
                \multicolumn{2}{c}{Num. Episodes}&-&2,226&4,452&6,678&8,904&11,131\\
                \multicolumn{2}{c}{Effective Epochs}&-&4.93&2.52&1.66&1.25&1.00\\
                \hline

\multirow{4}{*}{HF $\mathrm{wRMSE}$}
&LP(A)&\multirow{2}{*}{\textbf{0.530} $\pm$ 0.012}&0.555 $\pm$ 0.003 (+4.7\%)&0.554 $\pm$ 0.003 (+4.5\%)&0.637 $\pm$ 0.016 (+20.2\%)&0.614 $\pm$ 0.034 (+15.8\%)&0.613 $\pm$ 0.051 (+15.7\%)\\
&FT(A)&&0.543 $\pm$ 0.007 (+2.5\%)&\underline{0.535} $\pm$ 0.004 (+0.9\%)&0.555 $\pm$ 0.016 (+4.7\%)&0.548 $\pm$ 0.010 (+3.4\%)&0.547 $\pm$ 0.014 (+3.2\%)\\
\cline{2-8}
&LP(R)&\multirow{2}{*}{\textbf{0.501} $\pm$ 0.001}&0.561 $\pm$ 0.009 (+12.0\%)&0.574 $\pm$ 0.013 (+14.6\%)&0.563 $\pm$ 0.002 (+12.4\%)&0.578 $\pm$ 0.003 (+15.4\%)&0.567 $\pm$ 0.003 (+13.2\%)\\
&FT(R)&&0.506 $\pm$ 0.010 (+1.0\%)&0.506 $\pm$ 0.002 (+1.0\%)&\underline{0.503} $\pm$ 0.006 (+0.4\%)&0.513 $\pm$ 0.012 (+2.4\%)&0.513 $\pm$ 0.015 (+2.4\%)\\
\hline\hline

\multirow{4}{*}{LF $\mathrm{pRMSE}$}
&LP(A)&\multirow{2}{*}{0.511 $\pm$ 0.004}&0.528 $\pm$ 0.003 (+3.3\%)&0.530 $\pm$ 0.005 (+3.7\%)&0.579 $\pm$ 0.001 (+13.3\%)&0.574 $\pm$ 0.008 (+12.3\%)&0.568 $\pm$ 0.027 (+11.2\%)\\
&FT(A)&&\textbf{0.504} $\pm$ 0.001 (-1.4\%)&\underline{0.507} $\pm$ 0.002 (-0.8\%)&0.572 $\pm$ 0.011 (+11.9\%)&0.572 $\pm$ 0.012 (+11.9\%)&0.548 $\pm$ 0.043 (+7.2\%)\\
\cline{2-8}
&LP(R)&\multirow{2}{*}{0.533 $\pm$ 0.006}&0.541 $\pm$ 0.003 (+1.5\%)&0.539 $\pm$ 0.018 (+1.1\%)&0.552 $\pm$ 0.002 (+3.6\%)&0.552 $\pm$ 0.006 (+3.6\%)&0.548 $\pm$ 0.003 (+2.8\%)\\
&FT(R)&&\underline{0.494} $\pm$ 0.013 (-7.3\%)&\underline{0.494} $\pm$ 0.009 (-7.3\%)&\textbf{0.490} $\pm$ 0.016 (-8.1\%)&0.504 $\pm$ 0.008 (-5.4\%)&0.499 $\pm$ 0.019 (-6.4\%)\\
\hline\hline

\multirow{4}{*}{$\mathrm{CRPS}$}
&LP(A)&\multirow{2}{*}{0.517 $\pm$ 0.004}&0.514 $\pm$ 0.003 (-0.6\%)&0.514 $\pm$ 0.004 (-0.6\%)&0.549 $\pm$ 0.001 (+6.2\%)&0.546 $\pm$ 0.004 (+5.6\%)&0.542 $\pm$ 0.015 (+4.8\%)\\
&FT(A)&&\underline{0.508} $\pm$ 0.002 (-1.7\%)&\textbf{0.507} $\pm$ 0.004 (-1.9\%)&0.549 $\pm$ 0.007 (+6.2\%)&0.549 $\pm$ 0.005 (+6.2\%)&0.538 $\pm$ 0.023 (+4.1\%)\\
\cline{2-8}
&LP(R)&\multirow{2}{*}{0.518 $\pm$ 0.004}&0.519 $\pm$ 0.000 (+0.2\%)&0.517 $\pm$ 0.008 (-0.2\%)&0.525 $\pm$ 0.001 (+1.4\%)&0.524 $\pm$ 0.004 (+1.2\%)&0.523 $\pm$ 0.001 (+1.0\%)\\
&FT(R)&&\underline{0.498} $\pm$ 0.005 (-3.9\%)&\underline{0.498} $\pm$ 0.004 (-3.9\%)&\textbf{0.497} $\pm$ 0.005 (-4.1\%)&0.506 $\pm$ 0.013 (-2.3\%)&0.503 $\pm$ 0.014 (-2.9\%)\\

                \hline
                \hline
            \end{tabular}
            
        \end{center}
    \end{minipage}
\end{sidewaystable*}

\begin{table*}[!tb]
    \caption{Properties of pretraining and downstream datasets, including band energy ratios of the wrench. `LF' and `HF' indicate bands in $f\leq 1$~Hz and $f>1$~Hz, respectively.}
    \label{tbl:datasets_comparison}
    \setlength{\tabcolsep}{15pt}
    \renewcommand{\arraystretch}{1.1}
    \footnotesize
    \begin{center}
        \begin{tabular}{ccc}
            \hline\hline
            Datasets                        & Pretraining (RH20T)              & Downstream      \\
            \hline
            HF $\boldsymbol{W}$ Energy & 9.839\%                   & \textbf{85.188\%} \\
            LF $\boldsymbol{W}$ Energy & \textbf{90.161\%}         & 14.812\%          \\
            \hline
            Task                     & Everyday Manipulation & Block~Grinding    \\
            Actuation                       & Electric                  & Hydraulic         \\
            Actuation Signal                & Motor Torque              & Differential Pressure    \\
            Drive Type & Revolute & Prismatic and Revolute \\
            End-effector                    & Parallel Gripper          & Rotary Grinder    \\
            Robot Embodiment                & 6- and 7-DoF Arms         & 6-DoF Arm         \\
            F/T Location & Wrist & Wrist \\
            Control Mode & Human Teleoperation & Human Teleoperation \\
            \hline\hline
        \end{tabular}
    \end{center}
\end{table*}

%% file: src/sections/conclusions.tex
\section{Conclusions}\label{sec:conclusions}
% Overview
In this work, we propose a novel Frequency-aware Decomposition Network (FDN) for sensorless multi-step-ahead wrench estimation in vibration-rich robotic contact. 
% Problem
In our robotic grinding setting, the target wrench exhibits substantial task-critical vibrations that are challenging to estimate, whereas the hydraulic manipulator system complicates explicit identification of its dynamics. 
% Methods - decomp-based modeling + freq-aware layers
To this end, the proposed model adopts decomposition-based asymmetric modeling that combines pointwise regression and a conditional distribution to represent the low- and high-frequency wrench components. The integrated frequency-aware layers impose the band decomposition prior on the outputs and adaptively enhance the informative frequency components of the input. 
% Result (performance)
Under the assumed constant time delays, FDN outperforms the data-driven baselines in the high-frequency band and maintains overall competitive accuracy in the low-frequency band, while the others suffer from the accuracy trade-off between the two bands.
% Result (formulation)
Furthermore, the adopted multi-step-ahead formulation does not sacrifice accuracy relative to instantaneous estimation, indicating its practical feasibility in the real robot setting.
% Ablation studies
Ablation studies show the effectiveness of our asymmetric modeling and support our original design choices. 
% Result (performance+ablation)
Notably, FDN produces estimates over a 1,000~ms-ahead horizon within 11~ms on a single CPU thread, while also improving high-frequency amplitude fidelity by up to 47\%.
% Transfer learning
An exploratory study on transferring the learned dynamics from an open-source everyday manipulation dataset to our hydraulic grinding setting enables an additional 8\% accuracy gain in the low-frequency band, while the high-frequency dynamics appear more domain-specific.

We believe that the proposed framework can be useful for sensorless wrench estimation on industrial robots under vibration-rich, rapidly interacting conditions. Nevertheless, the present study has several limitations. 
First, the evaluation is limited to a single downstream task using one hydraulic manipulator. Collecting comparable vibration-rich interaction data requires substantial experimental resources, and future work will extend data collection and evaluation to other high-frequency-contact tasks such as drilling and cutting, as well as to different robot embodiments and tools. 
Second, although deployment constraints such as time delays and causal input treatment are accounted for, our evaluation is based on offline replay. Since FDN directly estimates the multi-step wrench horizon from causal proprioceptive inputs without recursive feedback, online deployment does not alter the estimator itself for a given input stream. Remaining deployment questions therefore mainly concern robustness to input distribution shifts and variable communication delays, as well as the practical utility of the estimated wrench in downstream real-time applications such as haptic feedback. Evaluating these aspects in an online setting lies beyond the scope of this estimation study and is left for future work.
Lastly, the transfer learning study is exploratory and covers a single source--target setting, where the transferability appears in the low-frequency band while the high-frequency dynamics remain domain-specific. Further studies on pretraining data selection, representation learning methods, and adaptation strategies would help extend transfer gains to the high-frequency band across a broader range of heterogeneous robot settings.

%% file: src/references.bib
@article{cheng2021mechanism,
  title     = {Mechanism-based structured deep neural network for cutting force forecasting using {CNC} inherent monitoring signals},
  author    = {Cheng, Yinghao and Li, Yingguang and Liu, Xu and Cai, Yu},
  journal   = {IEEE/ASME Transactions on Mechatronics},
  volume    = {27},
  number    = {4},
  pages     = {2235--2245},
  year      = {2022},
  publisher = {IEEE},
  doi = {10.1109/tmech.2021.3100719}
}

@article{ni2024unsupervised,
  title     = {Unsupervised domain adversarial adaptive regression network for cutting force prediction at varying spindle speeds},
  author    = {Ni, Chang and Yang, Jixiang and Ding, Han},
  journal   = {IEEE/ASME Transactions on Mechatronics},
  volume    = {30},
  number    = {1},
  pages     = {252--263},
  year      = {2025},
  publisher = {IEEE},
  doi = {10.1109/tmech.2024.3395436}
}

@article{dong2020contact,
  title     = {Contact force detection and control for robotic polishing based on joint torque sensors},
  author    = {Dong, Yunfei and Ren, Tianyu and Hu, Kui and Wu, Dan and Chen, Ken},
  journal   = {The International Journal of Advanced Manufacturing Technology},
  volume    = {107},
  number    = {5},
  pages     = {2745--2756},
  year      = {2020},
  publisher = {Springer},
  doi = {10.1007/s00170-020-05162-8}
}

@article{olsson2010cost,
  title     = {Cost-efficient drilling using industrial robots with high-bandwidth force feedback},
  author    = {Olsson, Tomas and Haage, Mathias and Kihlman, Henrik and Johansson, Rolf and Nilsson, Klas and Robertsson, Anders and Bj{\"o}rkman, Mats and Isaksson, Robert and Ossbahr, Gilbert and Brog{\aa}rdh, Torgny},
  journal   = {Robotics and Computer-Integrated Manufacturing},
  volume    = {26},
  number    = {1},
  pages     = {24--38},
  year      = {2010},
  publisher = {Elsevier},
  doi = {10.1016/j.rcim.2009.01.002}
}

@article{hu2025accurate,
  title     = {Accurate milling force estimation and surgical state recognition in robot-assisted laminectomy},
  author    = {Hu, Junfei and Zhou, Ziqi and Xia, Guangming and Dai, Yu and Zhang, Jianxun and Yang, Guihe and Han, Xiaoguang and Jiang, Jile and Liu, Yajun},
  journal   = {Measurement},
  volume    = {253},
  pages     = {117673},
  year      = {2025},
  publisher = {Elsevier},
  doi = {10.1016/j.measurement.2025.117673}
}

@article{xu2025grinding,
  title     = {Grinding force estimation and control of grinding robot with variable impedance control strategy},
  author    = {Xu, Du and Yin, Lairong and Wang, Jun},
  journal   = {The International Journal of Advanced Manufacturing Technology},
  volume    = {137},
  number    = {3},
  pages     = {2011--2027},
  year      = {2025},
  publisher = {Springer},
  doi = {10.1007/s00170-025-15137-2}
}

@inproceedings{zitkovich2023rt,
  title        = {{RT}-2: Vision-language-action models transfer web knowledge to robotic control},
  author       = {Zitkovich, Brianna and Yu, Tianhe and Xu, Sichun and Xu, Peng and Xiao, Ted and Xia, Fei and Wu, Jialin and Wohlhart, Paul and Welker, Stefan and Wahid, Ayzaan and others},
  booktitle    = {Conference on Robot Learning},
  pages        = {2165--2183},
  year         = {2023},
  organization = {PMLR}
}

@article{sliwowski2025reassemble,
  title   = {{REASSEMBLE}: A multimodal dataset for contact-rich robotic assembly and disassembly},
  author  = {Sliwowski, Daniel and Jadav, Shail and Stanovcic, Sergej and Orbik, Jedrzej and Heidersberger, Johannes and Lee, Dongheui},
  journal = {Robotics: Science and Systems XXI},
  year    = {2025},
  doi = {10.15607/rss.2025.xxi.059}
}

@article{wu2010overview,
  title     = {An overview of dynamic parameter identification of robots},
  author    = {Wu, Jun and Wang, Jinsong and You, Zheng},
  journal   = {Robotics and Computer-Integrated Manufacturing},
  volume    = {26},
  number    = {5},
  pages     = {414--419},
  year      = {2010},
  publisher = {Elsevier},
  doi = {10.1016/j.rcim.2010.03.013}
}

@inproceedings{eom1998disturbance,
  title        = {Disturbance observer based force control of robot manipulator without force sensor},
  author       = {Eom, Kwang Sik and Suh, Il Hong and Chung, Wan Kyun and Oh, S-R},
  booktitle    = {Proceedings. 1998 IEEE international conference on robotics and automation (Cat. No. 98CH36146)},
  volume       = {4},
  pages        = {3012--3017},
  year         = {1998},
  organization = {IEEE},
  doi = {10.1109/robot.1998.680888}
}

@article{tholey2005force,
  title     = {Force feedback plays a significant role in minimally invasive surgery: results and analysis},
  author    = {Tholey, Gregory and Desai, Jaydev P and Castellanos, Andres E},
  journal   = {Annals of Surgery},
  volume    = {241},
  number    = {1},
  pages     = {102--109},
  year      = {2005},
  publisher = {LWW},
  doi = {10.1097/01.sla.0000149301.60553.1e}
}

@inproceedings{stepputtis2022system,
  title        = {A system for imitation learning of contact-rich bimanual manipulation policies},
  author       = {Stepputtis, Simon and Bandari, Maryam and Schaal, Stefan and Amor, Heni Ben},
  booktitle    = {2022 IEEE/RSJ International Conference on Intelligent Robots and Systems (IROS)},
  pages        = {11810--11817},
  year         = {2022},
  organization = {IEEE},
  doi = {10.1109/iros47612.2022.9981802}
}

@article{muscolo2023force,
  title     = {Force--torque sensors for minimally invasive surgery robotic tools: An overview},
  author    = {Muscolo, Giovanni Gerardo and Fiorini, Paolo},
  journal   = {IEEE Transactions on Medical Robotics and Bionics},
  volume    = {5},
  number    = {3},
  pages     = {458--471},
  year      = {2023},
  publisher = {IEEE},
  doi = {10.1109/tmrb.2023.3261102}
}

@article{kao2020exploring,
  title     = {Exploring a Long Short-Term Memory based Encoder-Decoder framework for multi-step-ahead flood forecasting},
  author    = {Kao, I-Feng and Zhou, Yanlai and Chang, Li-Chiu and Chang, Fi-John},
  journal   = {Journal of Hydrology},
  volume    = {583},
  pages     = {124631},
  year      = {2020},
  publisher = {Elsevier},
  doi = {10.1016/j.jhydrol.2020.124631}
}

@inproceedings{kruvzic2022neural,
  title        = {Neural network-based end-effector force estimation for mobile manipulator on simulated uneven surfaces},
  author       = {Kru{\v{z}}i{\'c}, Stanko and Musi{\'c}, Josip and Stan{\v{c}}i{\'c}, Ivo and Papi{\'c}, Vladan},
  booktitle    = {2022 International Conference on Software, Telecommunications and Computer Networks (SoftCOM)},
  pages        = {1--6},
  year         = {2022},
  organization = {IEEE},
  doi = {10.23919/softcom55329.2022.9911383}
}

@article{pan2023adaptive,
  title     = {An adaptive sparse general regression neural network-based force observer for teleoperation system},
  author    = {Pan, Mingzhang and Li, Jing and Yang, Qiye and Wang, Yupeng and Tang, Yu and Pan, Lei and Jiang, Xianbao and Lin, Yizhong and Liang, Ke},
  journal   = {Engineering Applications of Artificial Intelligence},
  volume    = {118},
  pages     = {105689},
  year      = {2023},
  publisher = {Elsevier},
  doi = {10.1016/j.engappai.2022.105689}
}

@article{smith2006neural,
  title     = {Neural-network-based contact force observers for haptic applications},
  author    = {Smith, Andrew C and Mobasser, Farid and Hashtrudi-Zaad, Keyvan},
  journal   = {IEEE Transactions on Robotics},
  volume    = {22},
  number    = {6},
  pages     = {1163--1175},
  year      = {2006},
  publisher = {IEEE},
  doi = {10.1109/tro.2006.882923}
}

@article{dong2020sensorless,
  title     = {A sensorless interaction forces estimator for bilateral teleoperation system based on online sparse {Gaussian} process regression},
  author    = {Dong, Ai and Du, Zhijiang and Yan, Zhiyuan},
  journal   = {Mechanism and Machine Theory},
  volume    = {143},
  pages     = {103620},
  year      = {2020},
  publisher = {Elsevier},
  doi = {10.1016/j.mechmachtheory.2019.103620}
}

@article{pan2024graph,
  title     = {A graph robot network for force observer of teleoperation systems},
  author    = {Pan, Ming-Zhang and Li, Jing-Ao and Li, Zhen and Liang, Kun and Su, Tie-Cheng and Liang, Ke and Bian, Gui-Bin},
  journal   = {IEEE/ASME Transactions on Mechatronics},
  volume    = {30},
  number    = {1},
  pages     = {530--540},
  year      = {2025},
  publisher = {IEEE},
  doi = {10.1109/tmech.2024.3402995}
}

@article{chen2019rbf,
  title     = {{RBF}-neural-network-based adaptive robust control for nonlinear bilateral teleoperation manipulators with uncertainty and time delay},
  author    = {Chen, Zheng and Huang, Fanghao and Sun, Weichao and Gu, Jason and Yao, Bin},
  journal   = {{IEEE/ASME} Transactions on Mechatronics},
  volume    = {25},
  number    = {2},
  pages     = {906--918},
  year      = {2020},
  publisher = {IEEE},
  doi = {10.1109/tmech.2019.2962081}
}

@article{lee2024cnn,
  title     = {{cNN-DP}: Composite neural network with differential propagation for impulsive nonlinear dynamics},
  author    = {Lee, Hyeonbeen and Han, Seongji and Choi, Hee-Sun and Kim, Jin-Gyun},
  journal   = {Journal of Computational Physics},
  volume    = {496},
  pages     = {112578},
  year      = {2024},
  publisher = {Elsevier},
  doi = {10.1016/j.jcp.2023.112578}
}

@article{han2021dnn,
  title     = {A {DNN}-based data-driven modeling employing coarse sample data for real-time flexible multibody dynamics simulations},
  author    = {Han, Seongji and Choi, Hee-Sun and Choi, Juhwan and Choi, Jin Hwan and Kim, Jin-Gyun},
  journal   = {Computer Methods in Applied Mechanics and Engineering},
  volume    = {373},
  pages     = {113480},
  year      = {2021},
  publisher = {Elsevier},
  doi = {10.1016/j.cma.2020.113480}
}

@inproceedings{fang2024rh20t,
  title        = {{RH20T}: A comprehensive robotic dataset for learning diverse skills in one-shot},
  author       = {Fang, Hao-Shu and Fang, Hongjie and Tang, Zhenyu and Liu, Jirong and Wang, Chenxi and Wang, Junbo and Zhu, Haoyi and Lu, Cewu},
  booktitle    = {2024 IEEE International Conference on Robotics and Automation (ICRA)},
  pages        = {653--660},
  year         = {2024},
  organization = {IEEE},
  doi = {10.1109/icra57147.2024.10611615}
}

@article{butterworth1930theory,
  title   = {On the theory of filter amplifiers},
  author  = {Butterworth, Stephen},
  journal = {Wireless Engineer},
  volume  = {7},
  number  = {6},
  pages   = {536--541},
  year    = {1930}
}

@inproceedings{bahdanau2014neural,
  title     = {Neural machine translation by jointly learning to align and translate},
  author    = {Bahdanau, Dzmitry and Cho, Kyunghyun and Bengio, Yoshua},
  booktitle = {International Conference on Learning Representations},
  year      = {2015}
  }

@article{vaswani2017attention,
  title   = {Attention is all you need},
  author  = {Vaswani, Ashish and Shazeer, Noam and Parmar, Niki and Uszkoreit, Jakob and Jones, Llion and Gomez, Aidan N and Kaiser, {\L}ukasz and Polosukhin, Illia},
  journal = {Advances in Neural Information Processing Systems},
  volume  = {30},
  year    = {2017}
}

@inproceedings{zhou2021informer,
  title     = {Informer: Beyond efficient transformer for long sequence time-series forecasting},
  author    = {Zhou, Haoyi and Zhang, Shanghang and Peng, Jieqi and Zhang, Shuai and Li, Jianxin and Xiong, Hui and Zhang, Wancai},
  booktitle = {Proceedings of the AAAI Conference on Artificial Intelligence},
  volume    = {35},
  number    = {12},
  pages     = {11106--11115},
  year      = {2021},
  doi = {10.1609/aaai.v35i12.17325}
}

@article{wu2021autoformer,
  title   = {Autoformer: Decomposition transformers with auto-correlation for long-term series forecasting},
  author  = {Wu, Haixu and Xu, Jiehui and Wang, Jianmin and Long, Mingsheng},
  journal = {Advances in Neural Information Processing Systems},
  volume  = {34},
  pages   = {22419--22430},
  year    = {2021}
}

@inproceedings{zhou2022fedformer,
  title        = {{FEDformer}: Frequency enhanced decomposed transformer for long-term series forecasting},
  author       = {Zhou, Tian and Ma, Ziqing and Wen, Qingsong and Wang, Xue and Sun, Liang and Jin, Rong},
  booktitle    = {International Conference on Machine Learning},
  pages        = {27268--27286},
  year         = {2022},
  organization = {PMLR}
}

@inproceedings{nie2022time,
  title     = {A Time Series is Worth 64 Words: Long-term Forecasting with Transformers},
  author    = {Nie, Yuqi and Nguyen, Nam H. and Sinthong, Phanwadee and Kalagnanam, Jayant},
  booktitle = {International Conference on Learning Representations},
  year      = {2023}
  }

@inproceedings{kim2021reversible,
  title     = {Reversible instance normalization for accurate time-series forecasting against distribution shift},
  author    = {Kim, Taesung and Kim, Jinhee and Tae, Yunwon and Park, Cheonbok and Choi, Jang-Ho and Choo, Jaegul},
  booktitle = {International Conference on Learning Representations},
  year      = {2022}
}

@inproceedings{liu2023itransformer,
  title     = {{iTransformer}: Inverted transformers are effective for time series forecasting},
  author    = {Liu, Yong and Hu, Tengge and Zhang, Haoran and Wu, Haixu and Wang, Shiyu and Ma, Lintao and Long, Mingsheng},
  booktitle = {International Conference on Learning Representations},
  year      = {2024}
  }

@article{wang2024timexer,
  title   = {{TimeXer}: Empowering transformers for time series forecasting with exogenous variables},
  author  = {Wang, Yuxuan and Wu, Haixu and Dong, Jiaxiang and Qin, Guo and Zhang, Haoran and Liu, Yong and Qiu, Yunzhong and Wang, Jianmin and Long, Mingsheng},
  journal = {Advances in Neural Information Processing Systems},
  year    = {2024},
  doi = {10.52202/079017-0015}
}

@inproceedings{tayal2024exotst,
  title        = {{ExoTST}: Exogenous-Aware Temporal Sequence Transformer for Time Series Prediction},
  author       = {Tayal, Kshitij and Renganathan, Arvind and Jia, Xiaowei and Kumar, Vipin and Lu, Dan},
  booktitle    = {2024 IEEE International Conference on Data Mining (ICDM)},
  pages        = {857--862},
  year         = {2024},
  organization = {IEEE},
  doi = {10.1109/icdm59182.2024.00105}
}

@article{lim2021temporal,
  title     = {Temporal fusion transformers for interpretable multi-horizon time series forecasting},
  author    = {Lim, Bryan and Ar{\i}k, Sercan {\"O} and Loeff, Nicolas and Pfister, Tomas},
  journal   = {International Journal of Forecasting},
  volume    = {37},
  number    = {4},
  pages     = {1748--1764},
  year      = {2021},
  publisher = {Elsevier},
  doi = {10.1016/j.ijforecast.2021.03.012}
}

@inproceedings{rahaman2019spectral,
  title        = {On the spectral bias of neural networks},
  author       = {Rahaman, Nasim and Baratin, Aristide and Arpit, Devansh and Draxler, Felix and Lin, Min and Hamprecht, Fred and Bengio, Yoshua and Courville, Aaron},
  booktitle    = {International Conference on Machine Learning},
  pages        = {5301--5310},
  year         = {2019},
  organization = {PMLR}
}

@article{salinas2020deepar,
  title     = {{DeepAR}: Probabilistic forecasting with autoregressive recurrent networks},
  author    = {Salinas, David and Flunkert, Valentin and Gasthaus, Jan and Januschowski, Tim},
  journal   = {International Journal of Forecasting},
  volume    = {36},
  number    = {3},
  pages     = {1181--1191},
  year      = {2020},
  publisher = {Elsevier},
  doi = {10.1016/j.ijforecast.2019.07.001}
}

@article{tancik2020fourier,
  title   = {Fourier features let networks learn high frequency functions in low dimensional domains},
  author  = {Tancik, Matthew and Srinivasan, Pratul and Mildenhall, Ben and Fridovich-Keil, Sara and Raghavan, Nithin and Singhal, Utkarsh and Ramamoorthi, Ravi and Barron, Jonathan and Ng, Ren},
  journal = {Advances in Neural Information Processing Systems},
  volume  = {33},
  pages   = {7537--7547},
  year    = {2020}
}

@article{jacobs1991adaptive,
  title     = {Adaptive mixtures of local experts},
  author    = {Jacobs, Robert A and Jordan, Michael I and Nowlan, Steven J and Hinton, Geoffrey E},
  journal   = {Neural Computation},
  volume    = {3},
  number    = {1},
  pages     = {79--87},
  year      = {1991},
  publisher = {MIT Press},
  doi = {10.1162/neco.1991.3.1.79}
}

@article{hendrycks2016gaussian,
  title   = {Gaussian Error Linear Units ({GELUs})},
  author  = {Hendrycks, Dan and Gimpel, Kevin},
  journal = {arXiv preprint arXiv:1606.08415},
  year    = {2016}
  }

@inproceedings{kingma2014adam,
  title     = {Adam: A method for stochastic optimization},
  author    = {Kingma, Diederik P and Ba, Jimmy},
  booktitle = {International Conference on Learning Representations},
  year      = {2015}
  }

@article{savitzky1964smoothing,
  title     = {Smoothing and differentiation of data by simplified least squares procedures.},
  author    = {Savitzky, Abraham and Golay, Marcel JE},
  journal   = {Analytical Chemistry},
  volume    = {36},
  number    = {8},
  pages     = {1627--1639},
  year      = {1964},
  publisher = {ACS Publications},
  doi = {10.1021/ac60214a047}
}

@article{han2023multi,
  title     = {Multi-body dynamics model for spent nuclear fuel transportation system under normal transport test conditions},
  author    = {Han, Seongji and Jeong, Gil-Eon and Lee, Hyeonbeen and Choi, Woo-Seok and Kim, Jin-Gyun},
  journal   = {Nuclear Engineering and Technology},
  volume    = {55},
  number    = {11},
  pages     = {4125--4133},
  year      = {2023},
  publisher = {Elsevier},
  doi = {10.1016/j.net.2023.07.033}
}

@article{xu2025overview,
  title = {Overview {{Frequency Principle}}/{{Spectral Bias}} in {{Deep Learning}}},
  author = {Xu, Zhi-Qin John and Zhang, Yaoyu and Luo, Tao},
  year = 2025,
  journal = {Communications on Applied Mathematics and Computation},
  volume = {7},
  number = {3},
  pages = {827--864},
  doi = {10.1007/s42967-024-00398-7}
}

@article{werner2008estimation,
  title={On estimation of covariance matrices with {Kronecker} product structure},
  author={Werner, Karl and Jansson, Magnus and Stoica, Petre},
  journal={IEEE Transactions on Signal Processing},
  volume={56},
  number={2},
  pages={478--491},
  year={2008},
  publisher={IEEE},
  doi = {10.1109/tsp.2007.907834}
}

@article{ji2019industrial,
  title={Industrial robotic machining: a review},
  author={Ji, Wei and Wang, Lihui},
  journal={The International Journal of Advanced Manufacturing Technology},
  volume={103},
  number={1},
  pages={1239--1255},
  year={2019},
  publisher={Springer},
  doi = {10.1007/s00170-019-03403-z}
}

@incollection{han2024digital,
  title={Digital twin model of underwater construction robot for real-time grinding simulation},
  author={Han, Je Hyeop and Han, Jong-Boo and Kim, Sung-Soo and Kim, Myoung-Ho and Kim, Yong Hwan and Lee, Hyeonbeen and Kim, Jin-Gyun and Yeu, Tae-Kyeong},
  booktitle={IMSD2024},
  year={2024}
}

@article{park2025development,
  title={Development of human-in-the-loop based robot environment awareness and control framework for underwater robot operation},
  author={Park, Daegil and Han, Jong Boo and Lee, Yeongjun and Yeu, Tae-Kyeong},
  journal={Journal of Mechanical Science and Technology},
  volume={39},
  number={5},
  pages={2355--2363},
  year={2025},
  publisher={Springer},
  doi = {10.1007/s12206-025-2202-1}
}

@article{kim2019multi,
  title={Multi-axis force-torque sensors for measuring zero-moment point in humanoid robots: A review},
  author={Kim, Jung-Hoon},
  journal={IEEE Sensors Journal},
  volume={20},
  number={3},
  pages={1126--1141},
  year={2020},
  publisher={IEEE},
  doi = {10.1109/jsen.2019.2947719}
}

@article{norberto2002force,
  title={Force/torque sensing applied to industrial robotic deburring},
  author={Norberto Pires, J and Ramming, John and Rauch, Stephen and Ara{\'u}jo, Ricardo},
  journal={Sensor Review},
  volume={22},
  number={3},
  pages={232--241},
  year={2002},
  publisher={MCB UP Ltd},
  doi = {10.1108/02602280210433070}
}

@article{cao2021sixaxis,
  title = {Six-{{Axis Force}}/{{Torque Sensors}} for {{Robotics Applications}}: {{A Review}}},
  shorttitle = {Six-{{Axis Force}}/{{Torque Sensors}} for {{Robotics Applications}}},
  author = {Cao, Max Yiye and Laws, Stephen and y Baena, Ferdinando Rodriguez},
  year = 2021,
  journal = {IEEE Sensors Journal},
  volume = {21},
  number = {24},
  pages = {27238--27251},
  doi = {10.1109/JSEN.2021.3123638}
}

@article{ubeda2018design,
  title = {Design and {{Manufacturing}} of an {{Ultra-Low-Cost Custom Torque Sensor}} for {{Robotics}}},
  author = {Ubeda, Rodrigo Pérez and Gutiérrez Rubert, Santiago C. and Zotovic Stanisic, Ranko and Perles Ivars, {\'A}ngel},
  date = {2018-06-01},
  year=2018,
  journaltitle = {Sensors},
  shortjournal = {Sensors},
  volume = {18},
  number = {6},
  pages = {1786},
  doi = {10.3390/s18061786}
}

@article{choi2025coinft,
  title={{{CoinFT}}: A coin-sized, capacitive 6-axis force torque sensor for robotic applications},
  author={Choi, Hojung and Low, Jun En and Huh, Tae Myung and Hong, Seongheon and Uribe, Gabriela A and Hoffmann, Kenneth AW and Di, Julia and Chen, Tony G and Stanley, Andrew A and Cutkosky, Mark R},
  journal={arXiv preprint arXiv:2503.19225},
  year={2025}
}

@article{bicchi1992criterion,
  title={A criterion for optimal design of multi-axis force sensors},
  author={Bicchi, Antonio},
  journal={Robotics and Autonomous Systems},
  volume={10},
  number={4},
  pages={269--286},
  year={1992},
  publisher={Elsevier},
  doi = {10.1016/0921-8890(92)90005-j}
}

@article{li2023design,
  title={Design and application of multidimensional force/torque sensors in surgical robots: A review},
  author={Li, Zhangyong and Li, Xi and Lin, Jinzhao and Pang, Yu and Yang, Dewei and Zhong, Lisha and Guo, Jinhong},
  journal={IEEE Sensors Journal},
  volume={23},
  number={12},
  pages={12441--12454},
  year={2023},
  publisher={IEEE},
  doi = {10.1109/jsen.2023.3270229}
}

@article{kim2025parameter,
  title={Compact and optimized optical six-axis force/torque sensor for legged robots},
  author={Kim, Hyun-Bin and Ham, Byeong-Il and Choi, Keun-Ha and Kim, Kyung-Soo},
  journal={IEEE Transactions on Instrumentation and Measurement},
  year={2025},
  doi = {10.1109/TIM.2025.3618428},
  volume = {74},
  pages = {1--12}
}

@article{kim2024compact,
  title={A compact six-axis force/torque sensor using photocouplers for impact robustness},
  author={Kim, Hyun-Bin and Choi, Keun-Ha and Kim, Kyung-Soo},
  journal={Review of Scientific Instruments},
  volume={95},
  number={4},
  year={2024},
  publisher={AIP Publishing},
  doi = {10.1063/5.0186378}
}

@article{palli2015performance,
  title={Performance and sealing material evaluation in 6-axis force-torque sensors for underwater robotics},
  author={Palli, Gianluca and Moriello, Lorenzo and Melchiorri, Claudio},
  journal={IFAC-PapersOnLine},
  volume={48},
  number={2},
  pages={177--182},
  year={2015},
  publisher={Elsevier},
  doi = {10.1016/j.ifacol.2015.06.029}
}

@article{sun2022design,
  title={Design, manufacture, test and experiment of six-axis force torque sensor for {Chinese} experimental module manipulator},
  author={Sun, Yongjun},
  journal={Sensors},
  volume={22},
  number={9},
  pages={3603},
  year={2022},
  publisher={MDPI},
  doi = {10.3390/s22093603}
}

@article{xiong2020six,
  title={Six-dimensional force/torque sensor based on fiber {Bragg} gratings with low coupling},
  author={Xiong, Li and Guo, Yongxing and Jiang, Guozhang and Zhou, Xinglin and Jiang, Lin and Liu, Honghai},
  journal={IEEE Transactions on Industrial Electronics},
  volume={68},
  number={5},
  pages={4079--4089},
  year={2021},
  publisher={IEEE},
  doi = {10.1109/tie.2020.2982107}
}

@article{han2024data,
  title={Data-driven friction force prediction model for hydraulic actuators using deep neural networks},
  author={Han, Seongji and Orzechowski, Grzegorz and Kim, Jin-Gyun and Mikkola, Aki},
  journal={Mechanism and Machine Theory},
  volume={192},
  pages={105545},
  year={2024},
  publisher={Elsevier},
  doi = {10.1016/j.mechmachtheory.2023.105545}
}

@article{yang2025force,
  title={Force model of robot bone grinding based on finite element analysis},
  author={Yang, Kai and Jia, Qingxuan and Feng, Chao and Huang, Juxiang and Chen, Gang and Yang, Zheng},
  journal={Measurement},
  volume={243},
  pages={116124},
  year={2025},
  publisher={Elsevier},
  doi = {10.1016/j.measurement.2024.116124}
}

@article{armanini2023soft,
  title={Soft robots modeling: A structured overview},
  author={Armanini, Costanza and Boyer, Fr{\'e}d{\'e}ric and Mathew, Anup Teejo and Duriez, Christian and Renda, Federico},
  journal={IEEE Transactions on Robotics},
  volume={39},
  number={3},
  pages={1728--1748},
  year={2023},
  publisher={IEEE},
  doi = {10.1109/tro.2022.3231360}
}

@article{huang2017vehicle,
  title={Vehicle-manipulator system dynamic modeling and control for underwater autonomous manipulation},
  author={Huang, Hai and Tang, Qirong and Li, Hongwei and Liang, Le and Li, Weipo and Pang, Yongjie},
  journal={Multibody System Dynamics},
  volume={41},
  number={2},
  pages={125--147},
  year={2017},
  publisher={Springer},
  doi = {10.1007/s11044-016-9538-3}
}

@article{han2026updating,
  title={Updating polynomial chaos basis for addressing long time evaluation of multibody systems with uncertainties},
  author={Han, Seok-Hee and Choi, Hee-Sun and Kim, Jin-Gyun},
  journal={Multibody System Dynamics},
  volume={66},
  number={1},
  pages={11--47},
  year={2026},
  publisher={Springer},
  doi = {10.1007/s11044-025-10086-w}
}

@inproceedings{zahn1999learning,
  title={Learning friction estimation for sensorless force/position control in industrial manipulators},
  author={Zahn, Volker and Maass, R and Dapper, M and Eckmiller, Rolf},
  booktitle={Proceedings 1999 IEEE International Conference on Robotics and Automation (Cat. No. 99CH36288C)},
  volume={4},
  pages={2780--2785},
  year={1999},
  organization={IEEE},
  doi = {10.1109/robot.1999.774018}
}

@article{hu2018contact,
  title={Contact force estimation for robot manipulator using semiparametric model and disturbance {Kalman} filter},
  author={Hu, Jin and Xiong, Rong},
  journal={IEEE Transactions on Industrial Electronics},
  volume={65},
  number={4},
  pages={3365--3375},
  year={2018},
  publisher={IEEE},
  doi = {10.1109/tie.2017.2748056}
}

@inproceedings{hacksel1994estimation,
  title = {Estimation of Environment Forces and Rigid-Body Velocities Using Observers},
  booktitle = {Proceedings of the 1994 {{IEEE International Conference}} on {{Robotics}} and {{Automation}}},
  author = {Hacksel, P.J. and Salcudean, S.E.},
  year = 1994,
  pages = {931-936 vol.2},
  doi = {10.1109/ROBOT.1994.351233}
}

@article{alcocer2003force,
  title = {Force Estimation and Control in Robot Manipulators},
  author = {Alcocer, A. and Robertsson, A. and Valera, A. and Johansson, R.},
  year = 2003,
  journal = {IFAC Proceedings Volumes},
  volume = {36},
  number = {17},
  pages = {55--60},
  doi = {10.1016/S1474-6670(17)33369-4},
  copyright = {https://www.elsevier.com/tdm/userlicense/1.0/}
}

@inproceedings{deluca2003actuator,
  title = {Actuator Failure Detection and Isolation Using Generalized Momenta},
  booktitle = {2003 {{IEEE International Conference}} on {{Robotics}} and {{Automation}} ({{Cat}}. {{No}}.{{03CH37422}})},
  author = {De Luca, A. and Mattone, R.},
  year = 2003,
  volume = {1},
  pages = {634-639 vol.1},
  doi = {10.1109/ROBOT.2003.1241665}
}

@inproceedings{deluca2005sensorless,
  title = {Sensorless {{Robot Collision Detection}} and {{Hybrid Force}}/{{Motion Control}}},
  booktitle = {Proceedings of the 2005 {{IEEE International Conference}} on {{Robotics}} and {{Automation}}},
  author = {{de Luca}, A. and Mattone, R.},
  year = 2005,
  pages = {999--1004},
  doi = {10.1109/ROBOT.2005.1570247}
}

@inproceedings{murakami1993force,
  title = {Force Sensorless Impedance Control by Disturbance Observer},
  booktitle = {Conference {{Record}} of the {{Power Conversion Conference}} - {{Yokohama}} 1993},
  author = {Murakami, T. and Nakamura, R. and Yu, F. and Ohnishi, K.},
  year = 1993,
  pages = {352--357},
  doi = {10.1109/PCCON.1993.264158}
}

@inproceedings{ting2006bayesian,
  title={A {Bayesian} Approach to Nonlinear Parameter Identification for Rigid Body Dynamics.},
  author={Ting, Jo-Anne and Mistry, Michael N and Peters, Jan and Schaal, Stefan and Nakanishi, Jun},
  booktitle={Robotics: Science and Systems},
  pages={32--39},
  year={2006},
  organization={Philadelphia, USA},
  doi = {10.15607/rss.2006.ii.032}
}

@inproceedings{wahrburg2015cartesian,
  title={Cartesian contact force estimation for robotic manipulators using {Kalman} filters and the generalized momentum},
  author={Wahrburg, Arne and Morara, Elena and Cesari, Gianluca and Matthias, Bj{\"o}rn and Ding, Hao},
  booktitle={2015 IEEE International Conference on Automation Science and Engineering (CASE)},
  pages={1230--1235},
  year={2015},
  organization={IEEE},
  doi = {10.1109/coase.2015.7294266}
}

@article{jung2006robust,
  title={Robust contact force estimation for robot manipulators in three-dimensional space},
  author={Jung, J and Lee, J and Huh, K},
  journal={Proceedings of the Institution of Mechanical Engineers, Part C: Journal of Mechanical Engineering Science},
  volume={220},
  number={9},
  pages={1317--1327},
  year={2006},
  publisher={SAGE Publications Sage UK: London, England},
  doi = {10.1243/09544062c09005}
}

@article{liu2021sensorless,
  title={Sensorless force estimation for industrial robots using disturbance observer and neural learning of friction approximation},
  author={Liu, Sichao and Wang, Lihui and Wang, Xi Vincent},
  journal={Robotics and Computer-Integrated Manufacturing},
  volume={71},
  pages={102168},
  year={2021},
  publisher={Elsevier},
  doi = {10.1016/j.rcim.2021.102168}
}

@article{cao2021contact,
  title={Contact force and torque sensing for serial manipulator based on an adaptive {Kalman} filter with variable time period},
  author={Cao, Feng and Docherty, Paul D and Ni, Siliang and Chen, XiaoQi},
  journal={Robotics and Computer-Integrated Manufacturing},
  volume={72},
  pages={102210},
  year={2021},
  publisher={Elsevier},
  doi = {10.1016/j.rcim.2021.102210}
}

@article{mitsantisuk2011estimation,
  title={Estimation of action/reaction forces for the bilateral control using {Kalman} filter},
  author={Mitsantisuk, Chowarit and Ohishi, Kiyoshi and Katsura, Seiichiro},
  journal={IEEE Transactions on Industrial Electronics},
  volume={59},
  number={11},
  pages={4383--4393},
  year={2012},
  publisher={IEEE},
  doi = {10.1109/tie.2011.2173092}
}

@article{li2023noninvasive,
  title = {Non-Invasive Milling Force Monitoring through Spindle Vibration with {{LSTM}} and {{DNN}} in {{CNC}} Machine Tools},
  author = {Li, Bingchu and Liu, Taikang and Liao, Jun and Feng, Chunhua and Yao, Licheng and Zhang, Jingyang},
  year = 2023,
  journal = {Measurement},
  volume = {210},
  pages = {112554},
  doi = {10.1016/j.measurement.2023.112554}
}

@article{zhu2025physicsguided,
  title = {Physics-{{Guided Deep Network}} for {{Milling Dynamics Prediction}}},
  author = {Zhu, Kunpeng and Li, Jun},
  year = 2025,
  journal = {Engineering},
  volume = {55},
  pages = {71--85},
  doi = {10.1016/j.eng.2024.09.027}
}

@article{sun2019bilateral,
  title = {Bilateral Telerobotic System Using {{Type-2}} Fuzzy Neural Network Based Moving Horizon Estimation Force Observer for Enhancement of Environmental Force Compliance and Human Perception},
  author = {Sun, Da and Liao, Qianfang and Stoyanov, Todor and Kiselev, Andrey and Loutfi, Amy},
  year = 2019,
  journal = {Automatica},
  volume = {106},
  pages = {358--373},
  doi = {10.1016/j.automatica.2019.04.033}
}

@article{azimifar2017improving,
  title = {Improving Teleoperation System Performance in the Presence of Estimated External Force},
  author = {Azimifar, F. and Abrishamkar, M. and Farzaneh, B. and Sarhan, A.A.D. and Amini, H.},
  year = 2017,
  journal = {Robotics and Computer-Integrated Manufacturing},
  volume = {46},
  pages = {86--93},
  doi = {10.1016/j.rcim.2016.12.004}
}

@article{narayan2020developing,
  title = {Developing a Novel Force Forecasting Technique for Early Prediction of Critical Events in Robotics},
  author = {Narayan, Meenakshi and Fey, Ann Majewicz},
  editor = {Zhao, Huichan},
  year = 2020,
  journal = {PLOS ONE},
  volume = {15},
  number = {5},
  pages = {e0230009},
  doi = {10.1371/journal.pone.0230009}
}

@article{zhang2021polishing,
  title = {A {{Polishing Robot Force Control System Based}} on {{Time Series Data}} in {{Industrial Internet}} of {{Things}}},
  author = {Zhang, Chen and Tang, Zhuo and Li, Kenli and Yang, Jianzhong and Yang, Li},
  year = 2021,
  journal = {ACM Trans. Internet Technol.},
  volume = {21},
  number = {2},
  pages = {34:1--34:22},
  doi = {10.1145/3419469}
}

@article{scibilia2024nonlinear,
  title={A nonlinear modeling framework for force estimation in human-robot interaction},
  author={Scibilia, Adriano and Pedrocchi, Nicola and Fortuna, Luigi},
  journal={IEEE Access},
  volume={12},
  pages={97257--97268},
  year={2024},
  publisher={IEEE},
  doi = {10.1109/access.2024.3426942}
}

@article{jang2025sid,
  title={{SID-Net}: machine learning based system identification framework for rigid and flexible multibody dynamics},
  author={Jang, Sung Il and Han, Seongji and Kim, Jin-Gyun and Choi, Juhwan and Rhim, Sungsoo and Choi, Jin Hwan},
  journal={Multibody System Dynamics},
  volume={64},
  number={3},
  pages={415--436},
  year={2025},
  publisher={Springer},
  doi = {10.1007/s11044-024-10043-z}
}

@inproceedings{lee2026data,
  title={Data-Driven Pose Estimation for Robotic Connector Assembly: A Transformer-Based Approach using Force/Torque Time-Series},
  author={Lee, Joonho and Jung, Minjae and Feng, Jirou and Choi, Junhwan and Kim, Hwisoo and Ahn, Jeongdo and Park, Dong Il and Kim, Jin-Gyun and Kim, Jung},
  booktitle={2026 23rd International Conference on Ubiquitous Robots (UR)},
  pages={194--198},
  year={2026},
  organization={IEEE},
  doi = {10.1109/ur69537.2026.11626690}
}

@article{casolaro2023deep,
  title = {Deep {{Learning}} for {{Time Series Forecasting}}: {{Advances}} and {{Open Problems}}},
  shorttitle = {Deep {{Learning}} for {{Time Series Forecasting}}},
  author = {Casolaro, Angelo and Capone, Vincenzo and Iannuzzo, Gennaro and Camastra, Francesco},
  year = 2023,
  journal = {Information},
  volume = {14},
  number = {11},
  pages = {598},
  doi = {10.3390/info14110598}
}

@article{de200625,
  title={25 years of time series forecasting},
  author={De Gooijer, Jan G and Hyndman, Rob J},
  journal={International Journal of Forecasting},
  volume={22},
  number={3},
  pages={443--473},
  year={2006},
  publisher={Elsevier},
  doi = {10.1016/j.ijforecast.2006.01.001}
}

@article{li2024deep,
  title = {Deep {{Learning Models}} for {{Time Series Forecasting}}: {{A Review}}},
  shorttitle = {Deep {{Learning Models}} for {{Time Series Forecasting}}},
  author = {Li, Wenxiang and Law, K. L. Eddie},
  year = 2024,
  journal = {IEEE Access},
  volume = {12},
  pages = {92306--92327},
  doi = {10.1109/ACCESS.2024.3422528}
}

@inproceedings{yi2025surveya,
  title = {A {{Survey}} on {{Deep Learning}} Based {{Time Series Analysis}} with {{Frequency Transformation}}},
  booktitle = {Proceedings of the 31st {{ACM SIGKDD Conference}} on {{Knowledge Discovery}} and {{Data Mining V}}.2},
  author = {Yi, Kun and Zhang, Qi and Fan, Wei and Cao, Longbing and Wang, Shoujin and He, Hui and Long, Guodong and Hu, Liang and Wen, Qingsong and Xiong, Hui},
  year = 2025,
  pages = {6206--6215},
  publisher = {ACM},
  address = {Toronto ON Canada},
  doi = {10.1145/3711896.3736571},
  isbn = {979-8-4007-1454-2}
}

@inproceedings{piao2024fredformer,
  title = {Fredformer: {{Frequency Debiased Transformer}} for {{Time Series Forecasting}}},
  shorttitle = {Fredformer},
  booktitle = {Proceedings of the 30th {{ACM SIGKDD Conference}} on {{Knowledge Discovery}} and {{Data Mining}}},
  author = {Piao, Xihao and Chen, Zheng and Murayama, Taichi and Matsubara, Yasuko and Sakurai, Yasushi},
  year = 2024,
  pages = {2400--2410},
  publisher = {ACM},
  address = {Barcelona Spain},
  doi = {10.1145/3637528.3671928},
  isbn = {979-8-4007-0490-1}
}

@misc{liu2025freqmoe,
  title = {{{FreqMoE}}: {{Enhancing Time Series Forecasting}} through {{Frequency Decomposition Mixture}} of {{Experts}}},
  shorttitle = {{{FreqMoE}}},
  author = {Liu, Ziqi},
  year = 2025,
  number = {arXiv:2501.15125},
  eprint = {2501.15125},
  primaryclass = {cs.LG},
  publisher = {arXiv},
  doi = {10.48550/arXiv.2501.15125},
  archiveprefix = {arXiv}
}

@article{yilmaz2022transfer,
  title = {Transfer of Learned Dynamics between Different Surgical Robots and Operative Configurations},
  author = {Yilmaz, Nural and Zhang, Jintan and Kazanzides, Peter and Tumerdem, Ugur},
  year = 2022,
  journal = {International Journal of Computer Assisted Radiology and Surgery},
  volume = {17},
  number = {5},
  pages = {903--910},
  doi = {10.1007/s11548-022-02601-7}
}

@article{mccabe2024multiple,
  title={Multiple physics pretraining for spatiotemporal surrogate models},
  author={McCabe, Michael and R{\'e}galdo-Saint Blancard, Bruno and Parker, Liam and Ohana, Ruben and Cranmer, Miles and Bietti, Alberto and Eickenberg, Michael and Golkar, Siavash and Krawezik, Geraud and Lanusse, Francois and others},
  journal={Advances in Neural Information Processing Systems},
  volume={37},
  pages={119301--119335},
  year={2024},
  doi = {10.52202/079017-3791}
}

@article{batatia2025foundation,
  title={A foundation model for atomistic materials chemistry},
  author={Batatia, Ilyes and Benner, Philipp and Chiang, Yuan and Elena, Alin M and Kov{\'a}cs, D{\'a}vid P and Riebesell, Janosh and Advincula, Xavier R and Asta, Mark and Avaylon, Matthew and Baldwin, William J and others},
  journal={The Journal of Chemical Physics},
  volume={163},
  number={18},
  year={2025},
  publisher={AIP Publishing},
  doi = {10.1063/5.0297006}
}

@article{yang2025few,
  title={A few-shot transfer learning method for nonlinear system load identification via approximate-transfer-function-driven pretraining},
  author={Yang, Fengfan and Luo, Yajun and Lin, Ruishen and Zhang, Yahong and Xie, Shilin},
  journal={Mechanical Systems and Signal Processing},
  volume={237},
  pages={113077},
  year={2025},
  publisher={Elsevier},
  doi = {10.1016/j.ymssp.2025.113077}
}

@article{jaquier2025transfer,
  title={Transfer learning in robotics: An upcoming breakthrough? {A} review of promises and challenges},
  author={Jaquier, No{\'e}mie and Welle, Michael C and Gams, Andrej and Yao, Kunpeng and Fichera, Bernardo and Billard, Aude and Ude, Ale{\v{s}} and Asfour, Tamim and Kragic, Danica},
  journal={The International Journal of Robotics Research},
  volume={44},
  number={3},
  pages={465--485},
  year={2025},
  publisher={SAGE Publications Sage UK: London, England},
  doi = {10.1177/02783649241273565}
}

@article{miller2024survey,
  title={A survey of deep learning and foundation models for time series forecasting},
  author={Miller, John A and Aldosari, Mohammed and Saeed, Farah and Barna, Nasid Habib and Rana, Subas and Arpinar, I Budak and Liu, Ninghao},
  journal={arXiv preprint arXiv:2401.13912},
  year={2024}
}

@inproceedings{kim2024openvla,
  title     = {{OpenVLA}: An open-source vision-language-action model},
  author    = {Kim, Moo Jin and Pertsch, Karl and Karamcheti, Siddharth and Xiao, Ted and Balakrishna, Ashwin and Nair, Suraj and Rafailov, Rafael and Foster, Ethan and Lam, Grace and Sanketi, Pannag and others},
  booktitle = {Conference on Robot Learning},
  year      = {2024}
  }

@article{zhang2025time,
  title={Time series analysis in frequency domain: A survey of open challenges, opportunities and benchmarks},
  author={Zhang, Qianru and Sun, Yuting and Wen, Honggang and Yang, Peng and Li, Xinzhu and Li, Ming and Lam, Kwok-Yan and Yiu, Siu-Ming and Yin, Hongzhi},
  journal={arXiv preprint arXiv:2504.07099},
  year={2025}
}

@article{hu2026physicsinformedb,
  title = {Physics-Informed Adaptive {{Kalman}} Filter for Contact Force Estimation in Industrial Robots Considering Model Uncertainty},
  author = {Hu, Hongbo and Ma, Jia and Lu, Yuhan and Zhuang, Chungang},
  year = 2026,
  journal = {Robotics and Computer-Integrated Manufacturing},
  volume = {101},
  pages = {103292},
  doi = {10.1016/j.rcim.2026.103292}
}

@article{mu2025novel,
  title = {A Novel Dynamic Observer-Based Contact Force Control Strategy in Robotic Grinding to Improve Blade Profile Accuracy},
  author = {Mu, Yilin and Zou, Lai and Wang, Ziling and Li, Heng and Yan, Shengbo and Wang, Wenxi},
  year = 2025,
  journal = {Robotics and Computer-Integrated Manufacturing},
  volume = {94},
  pages = {102966},
  doi = {10.1016/j.rcim.2025.102966}
}

@misc{lee_2026_23026021,
  author       = {Lee, Hyeonbeen and Jung, Min-Jae and Yeu, Tae-Kyeong and Han, Jong-Boo and Park, Daegil and Stepputtis, Simon and Kim, Jin-Gyun},
  title        = {Frequency-aware decomposition learning for sensorless wrench estimation in vibration-rich robotic contact},
  howpublished = {Zenodo, Version 1.0},
  note         = {[dataset]},
  year         = {2026},
  doi          = {10.5281/zenodo.23026021}
}

@book{hildebrand1987introduction,
  title={Introduction to numerical analysis},
  author={Hildebrand, Francis Begnaud},
  year={1987},
  publisher={Courier Corporation}
}
